\documentclass[letterpaper]{article} 
\usepackage{aaai2026}  
\usepackage{times}  
\usepackage{helvet}  
\usepackage{courier}  
\usepackage[hyphens]{url}  
\usepackage{graphicx} 
\usepackage{natbib}  
\usepackage{caption} 
\usepackage{algorithm}
\usepackage{algorithmic}

\usepackage{url}            
\usepackage{booktabs}       
\usepackage{amsfonts}       
\usepackage{nicefrac}    
\usepackage{microtype}     

\usepackage{bbding}
\usepackage{pifont}
\usepackage{amssymb}

\usepackage{bm}
\usepackage{comment}
\usepackage{amsmath}
\usepackage{enumerate} 
\usepackage{enumitem}
\usepackage{mathrsfs}
\usepackage{makeidx}      
\usepackage{graphicx}          
\usepackage{epsfig,latexsym}
\usepackage{psfrag}
\usepackage{makecell}	

\newcommand{\mf}{\mathbf}

\newcommand{\cmark}{\ding{51}} 
\newcommand{\xmark}{\ding{55}}

\usepackage{newfloat}
\usepackage{listings}
\DeclareCaptionStyle{ruled}{labelfont=normalfont,labelsep=colon,strut=off} 
\floatstyle{ruled}
\newfloat{listing}{tb}{lst}{}
\floatname{listing}{Listing}
\title{Bring Your Dreams to Life: Continual Text-to-Video Customization}
\author{
    Jiahua Dong\textsuperscript{\rm 1}\equalcontrib, Xudong Wang\textsuperscript{\rm 2,3}\equalcontrib, Wenqi Liang\textsuperscript{\rm 4}, Zongyan Han\textsuperscript{\rm 1}, Meng Cao\textsuperscript{\rm 1}, Duzhen Zhang\textsuperscript{\rm 1}, \\ Hanbin Zhao\textsuperscript{\rm 5,6}, Zhi Han\textsuperscript{\rm 2}\thanks{Corresponding author.}, Salman Khan\textsuperscript{\rm 1,7}, Fahad Shahbaz Khan\textsuperscript{\rm 1,8}
}
\affiliations{
    \textsuperscript{\rm 1}Mohamed bin Zayed University of Artificial Intelligence 
    \textsuperscript{\rm 2}State Key Laboratory of Robotics and Intelligent Systems, \\Shenyang Institute of Automation, Chinese Academy of Sciences
    \textsuperscript{\rm 3}University of Chinese Academy of Sciences \\
    \textsuperscript{\rm 4}University of Trento
    \textsuperscript{\rm 5}Zhejiang University
    \textsuperscript{\rm 6}Zhejiang Key Laboratory of Intelligent High Speed UAV Technology \\
    \textsuperscript{\rm 7}Australian National University
    \textsuperscript{\rm 8}Link\"{o}ping University

     \{dongjiahua1995, liangwenqi0123\}@gmail.com, \{wangxudong, hanzhi\}@sia.cn, \{zongyan.han, meng.cao, duzhen.zhang, salman.khan, fahad.khan\}@mbzuai.ac.ae, zhaohanbin@zju.edu.cn
     
}

\begin{document}

\maketitle

\begin{abstract}
Customized text-to-video generation (CTVG) has recently witnessed great progress in generating tailored videos from user-specific text. However, most CTVG methods assume that personalized concepts remain static and do not expand incrementally over time. Additionally, they struggle with forgetting and concept neglect when continuously learning new concepts, including subjects and motions. To resolve the above challenges, we develop a novel \underline{C}ontinual \underline{C}ustomized \underline{V}ideo \underline{D}iffusion (CCVD) model, which can continuously learn new concepts to generate videos across various text-to-video generation tasks by tackling forgetting and concept neglect. To address catastrophic forgetting, we introduce a concept-specific attribute retention module and a task-aware concept aggregation strategy. They can capture the unique characteristics and identities of old concepts during training, while combining all subject and motion adapters of old concepts based on their relevance during testing. Besides, to tackle concept neglect, we develop a controllable conditional synthesis to enhance regional features and align video contexts with user conditions, by incorporating layer-specific region attention-guided noise estimation. Extensive experimental comparisons demonstrate that our CCVD outperforms existing CTVG baselines on both the DreamVideo and Wan 2.1 backbones. 
The code and models are available at \url{https://github.com/JiahuaDong/CCVD}.

\end{abstract}

\section{Introduction}
Text-to-video diffusion models (TVDMs) \cite{yang2025cogvideox, 10376647, 10.1007/978-3-031-72946-1_19} employ iterative denoising processes to synthesize realistic and dynamic video content directly from textual descriptions. By leveraging large-scale datasets and powerful spatiotemporal architectures \cite{wang2024vidprom}, they generate high-quality and contextually aligned videos that accurately reflect the semantics of the input text \cite{10377796}. 
Despite significant advancements, TVDMs often struggle with fine-grained control over object appearance and motion patterns, limiting their applicability in personalized tasks \cite{10655252, 10657098, 10191565}. To address this limitation, customized text-to-video generation (CTVG) \cite{10656326, he2025cameractrl} has been introduced to personalize video synthesis based on user-specified concepts. This involves integrating personalized subject identities and motion patterns into generated videos via the design of subject and motion adapters \cite{10.1007/978-3-031-73024-5_20, chen2024disenstudio}.

\begin{figure*}[t]
\vspace{-6mm}
\centering
\includegraphics[width = 0.97\linewidth]
{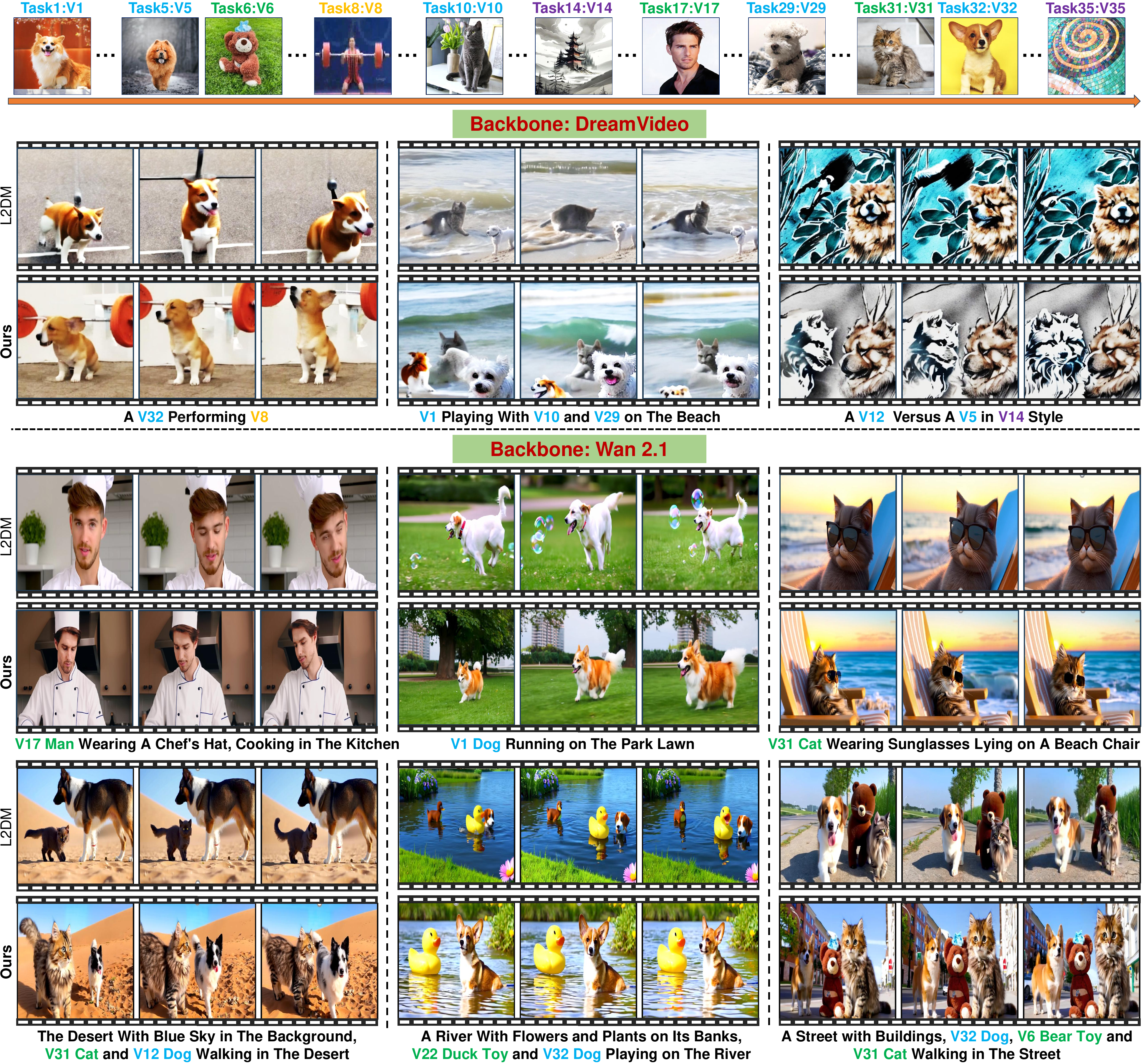}
\vspace{-3mm}
\caption{Results of our model in continually learning new personalized concepts, including subjects and motions, under the CTVC setting. Compared with L2DM, our model achieves better performance on both the DreamVideo and Wan 2.1 backbones.}
\label{fig: teaser}
\vspace{-3mm}
\end{figure*}

However, the majority of current CTVG methods \cite{10.1145/3641519.3657481, he2025cameractrl, she2025customvideox} are built on the premise that personalized concepts defined by users remain static and do not grow progressively over time. Nonetheless, this premise does not align with real-world scenarios, where users often seek to continuously create and synthesize video sequences with new personalized concepts (\emph{i.e.}, new subjects and motion patterns) based on their evolving experiences. To tackle this limitation, CTVG methods \cite{videomage, zhou2024sugarsubjectdrive, 10655226} generally need to retain all text-image training data associated with previous concepts to finetune pretrained TVDMs \cite{10657194} using the subject and motion adapters \cite{10656166}. Yet, the substantial computational expenses and privacy issues \cite{le_etal2023antidreambooth} can render them unfeasible as the quantity of personalized concepts gradually increases over time. If they preserve all subject and motion adapters tied to previously customized concepts and merge them to continuously adapt to new concepts \cite{wu2024mixture, zhong2024multi}, they face a substantial decline in the distinct characteristics of old concepts (\emph{i.e.}, catastrophic forgetting \cite{10555532952223295393, 8953627}). Besides, in practical applications, users typically seek to control the contexts (\emph{i.e.}, subjects and motions) linked to multiple old concepts in the generated videos by supplying additional conditions (\emph{e.g.}, bounding boxes). This imposes substantial strain on the above CTVG methods \cite{10656166, 10.1007/978-3-031-72992-8_16}, making them particularly vulnerable to the issue of concept neglect \cite{sun2024create} (\emph{i.e.}, certain old concepts may be overlooked in the multi-concept video customization).

To address the aforementioned real-world challenges, this paper introduces a novel practical problem referred to as \underline{C}ontinual \underline{T}ext-to-\underline{V}ideo \underline{C}ustomization (CTVC). 
Under the CTVC setting, as illustrated in Fig.~\ref{fig: teaser}, we can consecutively learn new personalized concepts (\emph{i.e.}, subjects and motion patterns) to generate video sequences across a variety of text-to-video generation tasks, including single/multi-concept video customization \cite{huang2025concevideo}, style transfer \cite{9093420} and video editing \cite{10377708}. Moreover, users can control the synthesis of video contexts, comprising both subjects and motion patterns, for multi-concept video customization by specifying certain conditions (\emph{e.g.}, bounding boxes). As previously mentioned, there are two challenges to conduct various text-to-video synthesis tasks in CTVC: \textbf{catastrophic forgetting} of previously learned concepts when continually learning new ones, and \textbf{concept neglect} when performing multi-concept video customization based on user-specified conditions.

To the CTVC problem, we propose a novel \underline{C}ontinual \underline{C}ustomized \underline{V}ideo \underline{D}iffusion (CCVD) model, which is the first work to tackle catastrophic forgetting and concept neglect across a variety of CTVG tasks.  
\textbf{First}, to overcome catastrophic forgetting of previously learned personalized concepts, we develop a concept-specific attribute retention (CAR) module for training and a task-aware concept aggregation (TCR) strategy for testing. The CAR module incorporates layer-specific concept tokens and a concept orthogonal loss to explore the unique characteristics and identity of personalized concepts, while the TCR module merges all subject and motion adapters of old concepts, weighted by their relevance to CTVG tasks. 
\textbf{Second}, we design a controllable conditional synthesis module to address concept neglect in the multi-concept video customization task. It employs layer-specific region attention to enhance the representational capacity of latent regional features and utilizes attention-guided noise estimation to control the context of personalized subjects and motions within the generated video, aligning with user-specified conditions. Experiments demonstrate the effectiveness of our CCVD model on both the DreamVideo and Wan 2.1 backbones. 
The key contributions are as follows:
\begin{itemize}
\item We introduce a new problem termed Continual Text-to-Video Customization (CTVC), characterized by two key challenges: catastrophic forgetting and concept neglect. 

\item We develop a Continual Customized Video Diffusion (CCVD) model, capable of continuously learning new concepts while tackling forgetting and concept neglect.

\item We propose a concept-specific attribute retention module and a task-aware concept aggregation strategy to mitigate catastrophic forgetting of previously learned concepts.

\item We devise a controllable conditional synthesis module to address concept neglect by leveraging layer-specific region attention and attention-guided noise estimation. 

\end{itemize}

\section{Related Work}
\textbf{Customized text-to-video generation} (CTVG) integrates user-specific subjects and motion patterns to synthesize personalized videos while maintaining contextual consistency and visual quality. Following previous works \cite{gal2023an, NEURIPS2024Dong}, which addresses subject-driven and style-driven concept customization by leveraging a few images to finetune Stable Diffusion \cite{10205187, wang2024instantas}, recent studies \cite{10656166, jiang2024videobooth, 10.1007/978-3-031-73024-5_20} have extended them into the temporal domain for customized text-to-video generation. DreamVideo \cite{10656166} introduces a two-stage learning framework that decouples the subject learning from motion learning. MotionBooth \cite{wu2025motionbooth} incorporates subject region loss and video preservation loss to enhance alignment between the input concept and generated visual content. Additionally, \cite{chen2024disenstudio, she2025customvideox, liang2025movie} have addressed the more complex challenge of multi-concept video customization. ConceptMaster \cite{huang2025concevideo} tackles multi-concept video customization via decoupled concept embeddings and systematic data construction. However, existing methods \cite{Skyreels, jiang2024videobooth} encounter the issues of catastrophic forgetting and concept neglect when continually learning new concepts for video customization. 

\textbf{Continual Learning} (CL) \cite{li2017learning, 9878745} enables models to incrementally acquire new knowledge while retaining previously learned information without retraining from scratch. Some regularization methods incorporate constraint terms to limit modifications to network weights associated with previous tasks \cite{Kirkpatrick3521, dong2022federated_FCIL}, or employ model distillation \cite{10.1007/978-3-319-46493-0_37, wei2024class} to transfer knowledge between previous and current models. Approaches based on data replay preserve samples from old categories \cite{9009019} or utilize generative models to produce synthetic data belonging to old classes \cite{qi2025classwise}. Structural adaptation methods \cite{yoon2018lifelong, Douillard_2022_CVPR} progressively modify their architectures to incorporate new information while maintaining previous knowledge. Besides, feature-based methods leverage contrastive mechanisms \cite{madaan2022representational} or prompt techniques \cite{masum2024prototypesinjected, Smith_2023_CVPR} to facilitate ongoing task acquisition. Yet, these approaches \cite{wei2025compress, dong_10323204} cannot be directly employed to address continual text-to-video customization problem.

\section{Preliminary and Problem Definition}
\textbf{Preliminary:}
Text-to-video diffusion models (TVDMs) \cite{ModelScopeT2V, 10203078, wang2023videocomposer, Wu_Zhang_Wangcdcd} use textual prompts as input to generate videos that visually match the content described in the prompts. They synthesize videos in the latent feature space by introducing an encoder $\mathcal{E}(\cdot)$ and a decoder $\mathcal{D}(\cdot)$. Customized text-to-video generation (CTVG) \cite{10656166, 1011453687945} uses a subject adapter and a motion adapter to generate customized videos by finetuning the pretrained TVDMs \cite{10655252, 10657098}. Given a personalized image $\mf{x}$ and its text prompt $\mf{p}$, the encoder $\mathcal{E}(\cdot)$ projects $\mf{x}$
to a latent feature $\mf{z}$, and the text encoder $\Phi(\cdot)$ encodes $\mf{p}$ to the textual embedding $\mf{c} = \Phi(\mf{p})$. Here, $\mf{x}$ is a personalized pattern. The objective to synthesize a custom video for the given concept $\{\mf{x}, \mf{p}\}$ is defined as: 
\begin{align}
\mathcal{L}_{\mathrm{CTVG}} = \mathbb{E}_{\mathbf{z}\sim\mathcal{E}(\mathbf{x}), \mathbf{c}, \epsilon\sim \mathcal{N}(0, \mathbf{I}), t}[\|\epsilon - \epsilon_{\theta} (\mathbf{z}_t, \mathbf{c}, t)\|_2^2],
\label{eq: CVDM_loss}
\end{align}
where $\mf{z}_t$ is the noisy latent feature at the $t$-th timestep. $\epsilon_{\theta}(\cdot)$ represents a 3D UNet \cite{ModelScopeT2V} that predicts the noise estimation $\epsilon_{\theta} (\mathbf{z}_t, \mathbf{c}, t)$, guiding Gaussian noise $\mathcal{N}(0, \mf{I})$.  

\begin{figure*}[t]
\vspace{-7mm}
\centering
\includegraphics[width = 1.0\linewidth]
{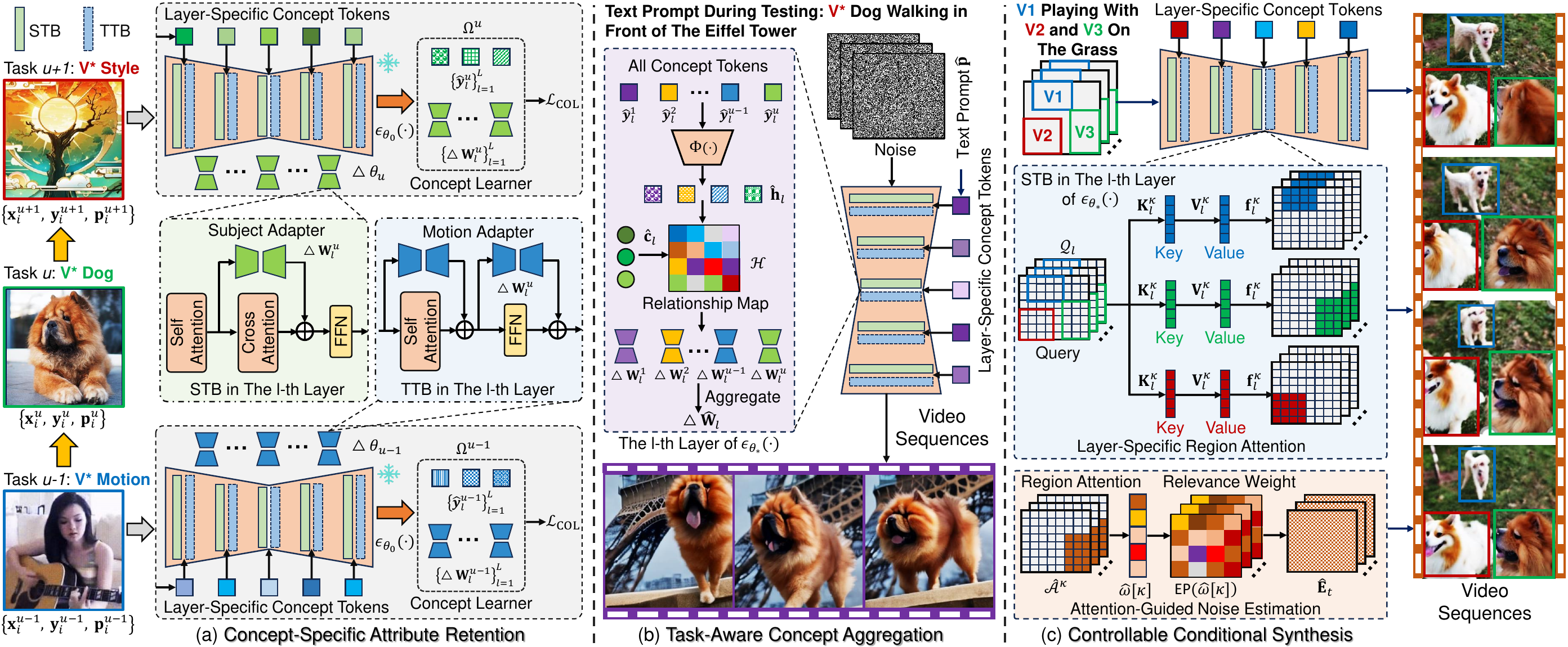}
\vspace{-7mm}
\caption{CCVD architectural overview. It includes (a) a concept-specific attribute retention module, (b) a task-aware concept aggregation strategy to overcome catastrophic forgetting of previous concepts during training and testing, and (c) a controllable conditional synthesis with layer-specific region attention and attention-guided noise estimation to address concept neglect.}
\label{fig: model_pipeline}
\end{figure*}

In $\epsilon_{\theta}(\cdot)$, $\theta = \theta_0+\triangle\theta$ and each layer contains a temporal transformer block (TTB) and a spatial transformer block (STB). Here, $\theta_0=\{\mf{W}_l^0\}_{l=1}^L$ denotes the parameters of pretrained TVDMs, and $\triangle\theta= \{\triangle\mf{W}_l\}_{l=1}^L$ indicates the parameters of the subject adapter or motion adapter. In the $l$-th ($l=1, \cdots, L$) layer of $\epsilon_{\theta}(\cdot)$, $\mf{W}_l^0 \in\mathbb{R}^{r\times s}$ is the pretrained weight, and $\triangle\mf{W}_l\in\mathbb{R}^{r\times s}$ denotes the weight of the subject adapter or motion adapter. $r$ and $s$ are the matrix dimension. As proposed in \cite{10656166}, $\triangle\mf{W}_l = \mf{R}_l \mf{S}_l$ consists of a down-projected layer $\mf{R}_l\in\mathbb{R}^{r\times b}$ and an up-projected layer $\mf{S}_l\in\mathbb{R}^{b\times s}$, where $b$ is the feature dimension. As shown in Fig.~\ref{fig: model_pipeline}(a), if $\mf{x}$ is a personalized subject, $\triangle\mf{W}_l$ is placed in the STB; otherwise, it is placed in the TTB to learn the motion pattern. However, most existing CTVG methods \cite{10656166, huang2025concevideo} presume that the number of personalized concepts (\emph{i.e.}, subjects and motion patterns) for each user stays fixed over time. This assumption does not hold in practical scenarios, where users typically want to continuously synthesize new personalized concepts according to their evolving experiences. Furthermore, they face the challenge of catastrophic forgetting on previously learned personalized concepts and experience concept neglect when synthesizing customized videos in concept-incremental way. 

\textbf{Problem Definition:}
For the above challenges, we propose a new Continual Text-to-Video Customization (CTVC) problem. Define $U$ continuous customized text-to-video generation tasks as $\mathcal{T}=\{\mathcal{T}^u\}_{u=1}^U$, where the $u$-th task $\mathcal{T}^u=\{\mf{x}_i^u, \mf{y}_i^u, \mf{p}_i^u\}_{i=1}^{n_u}$ comprises $n_u$ images $\mf{x}_i^u$, concept tokens $\mf{y}_i^u \in\mathcal{Y}^u$ and their corresponding text prompts $\mf{p}_i^u$. In this paper, $\mf{p}_i^u$ refers to the textual description of a subject or motion pattern (\emph{e.g.}, [$V_*$] [$V_{\mathrm{cat}}$] swims in the pool or a boy [$V_{\mathrm{plays}}$] football), and $\mf{y}_i^u$ indicates the concept token (\emph{e.g.}, [$V_*$] [$V_{\mathrm{cat}}$] or [$V_{\mathrm{plays}}$]) in $\mf{p}_i^u$. 
$\mathcal{Y}^u$ denotes the concept space of the $u$-th task, which consists of $K^u$ new concepts $\mathbf{y}^u = \cup_{i=1}^{n_u} \mathbf{y}_i^u$. In CTVC, the concept space $\mathcal{Y}^u$ of the $u$-th task does not overlap with any previous tasks: $\mathcal{Y}^u\cap(\cup_{j=1}^{u-1}\mathcal{Y}^j)=\emptyset$. It indicates that $K^u$ new concepts in $\mathcal{T}^u$ are different from $\sum_{j=1}^{u-1}K^j$ previously learned concepts from $\cup_{j=1}^{u-1}\mathcal{T}^j$. According to users' preference, the $u$-th task $\mathcal{T}^u$ may be one of the following versatile video customization tasks: single/multi-concept video synthesis \cite{huang2025concevideo}, video editing \cite{Feng_2024_CVPR}, and style transfer \cite{8237388}. Given practical and privacy concerns, we do not replay or store any data from old tasks, allowing us to learn all video customization tasks in a incremental way. CTVC aims to continually learn a sequence of concepts while alleviating forgetting and concept neglect of old concepts. 

\section{The Proposed Model}\label{sec:proposed_model}
The architecture of CCVD, designed to resolve the CTVC problem, is illustrated in Fig.~\ref{fig: model_pipeline}. It comprises three key components: (a) a concept-specific attribute retention, (b) a task-aware concept aggregation strategy to mitigate catastrophic forgetting, and (c) a controllable conditional synthesis module, which includes layer-specific region attention and attention-guided noise estimation to solve concept neglect.

\subsection{Concept-Specific Attribute Retention}\label{sec:concept_specific_attribute_retention}
Motivated by \cite{10656166}, if the $K^u$ new concepts in the $u$-th task $\mathcal{T}^u$ belong to subjects, we place the subject adapter in the spatial transformer block (STB) to finetune the pretrained 3D UNet $\epsilon_{\theta_0}(\cdot)$ via Eq.~\eqref{eq: CVDM_loss} for learning $\mathcal{T}^u$. Otherwise, we place the motion adapter in the temporal transformer block (TTB) to learn personalized motion patterns in $\mathcal{T}^u$. After learning the $u$-th ($u=1, \cdots, U$) task, we obtain an updated 3D UNet $\epsilon_{\theta_u}(\cdot)$, where $\theta_u = \theta_0+\triangle\theta_u$. Here, $\triangle\theta_u=\{\triangle\mf{W}_l^u\}_{l=1}^{L}$ denotes the updated parameters of the subject adapter or motion adapter of the $u$-th task. $\triangle\mf{W}_l^u = \mf{R}_l^u \mf{S}_l^u$ is the updated weight in the $l$-th layer of $\epsilon_{\theta_u}(\cdot)$, $\mf{R}_l^u\in\mathbb{R}^{r\times b}$ and $\mf{S}_l^u\in\mathbb{R}^{b\times s}$ are the down-projected and up-projected layers. Intuitively, $\triangle\theta_u$ is capable of capturing the majority of personalized concept identity in $\mathcal{T}^u$. In light of this, a straightforward approach to resolving the CTVC problem involves retaining all updated subject and motion adapters $\{\triangle\theta_j\}_{j=1}^{u-1}$ learned so far, and combining them linearly by assessing their respective contributions to learning personalized concepts. However, continually learning new concepts in the CTVC setting may cause this trivial approach to suffer a significant loss of unique attributes in certain personalized concepts. This issue is commonly termed catastrophic forgetting \cite{smith2023continual, sun2024create} of previously learned concepts. To alleviate catastrophic forgetting, as depicted in Fig.~\ref{fig: model_pipeline}(a), we propose a concept-specific attribute retention module. This module incorporates layer-specific concept tokens and a concept orthogonal loss to capture unique concept attributes within different tasks. 

\textbf{Layer-Specific Concept Tokens:}
Unlike DreamVideo \cite{10656166}, which uses textual inversion \cite{gal2023an, Zhang_2023_inst, voynov2023P} to apply a uniform text prompt across all layers, we propose layer-specific concept tokens to enhance the preservation of unique attributes associated with old concepts. At the $u$-th video customization task, $L$ layer-specific concept tokens $\{\mf{y}_{i,l}^u\}_{l=1}^L$ are defined for any triplet $\{\mf{x}_i^u, \mf{y}_i^u, \mf{p}_i^u\}\in\mathcal{T}^u$. Specifically, given a textual description $\mf{p}_{i}^u$ ([$V_*$] [$V_{\mathrm{cat}}$] swims in the pool), we denote the layer-specific text prompt as $\mf{p}_{i,l}^u$ ([$V_*^l$] [$V_{\mathrm{cat}}^l$] swims in the pool), and the layer-specific concept token as $\mf{y}_{i,l}^u$ ([$V_*^l$] [$V_{\mathrm{cat}}^l$]) at the $l$-th layer of $\epsilon_{\theta_u}(\cdot)$. Subsequently, we initialize $\{\mf{p}_{i,l}^u\}_{l=1}^L$ using $\mf{p}_i^u$ and integrate the textual embedding $\mf{c}_{i,l}^u = \Phi(\mf{p}_{i,l}^u)$, which is encoded through the text encoder $\Phi(\cdot)$, into the $l$-th layer of $\epsilon_{\theta_u}(\cdot)$. When we train $\epsilon_{\theta_u}(\cdot)$ via Eq.~\eqref{eq: CVDM_loss} to synthesize customized videos, the layer-specific concept tokens are effective to explore distinctive and inherent identity attributes of previously learned concepts, thereby overcoming the catastrophic forgetting.

\textbf{Concept Orthogonal Loss:}
Under the CTVC setting, the discriminative capacity of unique identity characteristics to differentiate between various personalized concepts may substantially degrade as the continual learning of a series of video customization tasks progresses, particularly for semantically similar concepts. To resolve this issue, we introduce a concept orthogonal loss $\mathcal{L}_{\mathrm{COL}}$ to perform orthogonality constrain among different concept subspaces across video customization tasks. 
Specifically, the updated weight $\triangle\mf{W}_l^u = \mf{R}_l^u \mf{S}_l^u \in\mathbb{R}^{r\times s}$ at the $l$-th layer of $\epsilon_{\theta_u}(\cdot)$ can be interpreted as comprising two key components: the concept subspace of the $u$-th task $\mathbf{R}_l^u = [\mathbf{r}_{l}^{u,1}, \cdots, \mathbf{r}_{l}^{u,b}]\in\mathbb{R}^{r\times b}$, and its corresponding weighting coefficient $\mathbf{S}^u_l = [\mathbf{s}_{l}^{u,1}, \cdots, \mathbf{s}_{l}^{u,b}]^\top \in\mathbb{R}^{b\times s}$. Here, $\mathbf{r}_{l}^{u,i}\in\mathbb{R}^r$ is the $i$-th ($i=1, \cdots, b$) basis vector of the concept space $\mathbf{R}_l^u$ and $\mathbf{s}_{l}^{u,i} \in\mathbb{R}^s$ denotes its linear coefficient. When learning the $u$-th video customization task $\mathcal{T}^u$, we want the concept subspace of the $u$-th task to be orthogonal to the concept subspaces of previous tasks $\cup_{j=1}^{u-1}\mathcal{T}^j$: $\sum_{j=1}^{u-1}\sum_{l=1}^{L} tr(\mathbf{R}_l^j (\mathbf{R}_l^u)^\top\!)=0$. 
We devise an alternative optimization approach aimed at minimizing the absolute value of the inner product among distinct subspaces. As a result, we formulate the concept orthogonal loss $\mathcal{L}_{\mathrm{COL}}$ as follows:
\begin{align}
\mathcal{L}_{\mathrm{COL}} =&\mathbb{E}_{\mathbf{z}\sim\mathcal{E}(\mathbf{x}), \mathbf{c}, \epsilon\sim \mathcal{N}(0, \mathbf{I}), t}\big[\|\epsilon - \epsilon_{\theta_u} (\mathbf{z}_t, \mathbf{c}, t)\|_2^2 + \nonumber \\ \label{eq: COL_loss}
&\qquad\qquad\quad \lambda\sum\nolimits_{j=1}^{u-1}\sum\nolimits_{l=1}^{L} tr(\mathbf{R}_l^j (\mathbf{R}_l^u)^\top) \big],
\end{align}
where $\lambda$ is a hyper-parameter. The loss $\mathcal{L}_{\mathrm{COL}}$ helps the layer-specific concept tokens better explore the distinctive attributes of previous concepts, especially for semantically similar ones, by reducing correlations among concept subspaces. Upon finishing the learning of the $u$-th video customization task via Eq.~\eqref{eq: COL_loss}, we obtain the task-wise concept learner $\Omega^u=\{\triangle\mathbf{W}_l^u, \widehat{\mathbf{y}}_l^u\}_{l=1}^L$, where $\widehat{\mathbf{y}}_l^u = \{\widehat{\mathbf{y}}_l^{u,k}\}_{k=1}^{K^u}$ and $\widehat{\mathbf{y}}_l^{u,k}$ is the $k$-th learned concept tokens at the $l$-th layer.

\subsection{Task-Aware Concept Aggregation}\label{sec:task_aware_concept_aggregation}
In order to alleviate catastrophic forgetting of previously learned concept during the testing phase, we retain all task-wise concept learners $\{\Omega^j\}_{j=1}^{u}$ learned so far, and propose a task-aware concept aggregation strategy to dynamically aggregate these concept learners for customized text-to-video generation tasks.     
As in Fig.~\ref{fig: model_pipeline}(b), for a text prompt $\widehat{\mf{p}}$ during testing, we first initialize the layer-specific text prompt $\{\widehat{\mathbf{p}}_l\}_{l=1}^L$ using $\widehat{\mf{p}}$, and then generate layer-specific textual embeddings $\widehat{\mathbf{c}} = \{\widehat{\mathbf{c}}_l\in\mathbb{R}^{n_e\times d}\}_{l=1}^L$
through the text encoder $\Phi(\cdot)$. Here, $n_e$ is the number of tokens and $d$ denotes the embedding dimension. Subsequently, we gather all concept tokens $\{\mathcal{Z}_l\}_{l=1}^L$ from stored concept learners $\{\Omega^j\}_{j=1}^{u}$, where $\mathcal{Z}_l= \cup_{j=1}^u \widehat{\mathbf{y}}_l^j \in\mathbb{R}^{n_t}$ comprises $n_t=\sum_j^{u}K^j$ layer-specific concept tokens at the $l$-th layer. After projecting $\{\mathcal{Z}_l\}_{l=1}^L$ to latent embeddings $\{\mf{h}_l \in\mathbb{R}^{n_t\times d}\}_{l=1}^L$ using $\Phi(\cdot)$, we compute the average of these latent embeddings corresponding to the same task, resulting in $\{\widehat{\mf{h}}_l \in\mathbb{R}^{u\times d}\}_{l=1}^L$. $u$ is the number of tasks learned up until now. After calculating the relationships $\mathcal{H}\in\mathbb{R}^{u}$ between $\widehat{\mf{h}}_l$ and $\widehat{\mathbf{c}}_l$, we leverage $\mathcal{H}$ to dynamically aggregate $\{\triangle\mf{W}_l^j\}_{j=1}^u$ at the $l$-th layer to obtain $\triangle\widehat{\mathbf{W}}_l$: 
\begin{align}
\!\!\mathcal{H} \!=\! \max\big( \mathbf{\widehat{c}}_l\cdot (\mathbf{\widehat{h}}_l)^\top \big),~\triangle\widehat{\mathbf{W}}_l \!=\! \sum\nolimits_{j=1}^u \!\triangle\mathbf{W}_l^j \cdot \zeta(\mathcal{H})_j,\!\!
\label{eq: concept_aggregation}
\end{align}
where $\max(\cdot)$ is the maximization along the row axis. The normalization of $\mathcal{H}$ is achieved by $\zeta(\mathcal{H}) = \mathcal{H}^2 / \|\mathcal{H}^2\|_F \in\mathbb{R}^u$, and $\zeta(\mathcal{H})_j$ corresponds to the $j$-th element of $\zeta(\mathcal{H})$.

\textbf{Test Model:}
After computing $\triangle\theta_*=\{\triangle\widehat{\mathbf{W}}_l\}_{l=1}^L$ via Eq.~\eqref{eq: concept_aggregation}, we derive a new 3D UNet $\epsilon_{\theta_*}(\cdot)$ for testing, where $\theta_*= \theta_0+\triangle\theta_*$. 
At the $l$-th layer of $\epsilon_{\theta_*}(\cdot)$, if the text prompt $\widehat{\mf{p}}$ belongs to a subject, we position $\triangle\widehat{\mathbf{W}}_l$ in the STB; if $\widehat{\mf{p}}$ describes a motion pattern, we place $\triangle\widehat{\mathbf{W}}_l$ in TTB; if $\widehat{\mf{p}}$ encompasses both subject and motion, we position $\triangle\widehat{\mathbf{W}}_l$ in STB and TTB respectively. Particularly, $\epsilon_{\theta_*}(\cdot)$ has incorporated significant unique attributes of all concepts learned up until now, effectively reducing catastrophic forgetting of previously learned concepts during the testing phase.

\subsection{Controllable Conditional Synthesis}\label{sec:controllable_conditional_synthesis}
When employing for multi-concept customization, it fails to produce high-fidelity videos that align with user-specified conditions (\emph{e.g.}, bounding boxes), due to concept neglect (\emph{i.e.}, some subjects or motions are omitted in the generated videos). To tackle this issue, as depicted in Fig.~\ref{fig: model_pipeline}(c), we develop a layer-specific region attention and an attention-guided noise estimation for customized multi-concept video generation. 

\textbf{Layer-Specific Region Attention:}
In addition to the initial text prompt $\widehat{\mf{p}}$, users can utilize $n_r$ regional conditions $\{\widehat{\mf{p}}^\kappa, \widehat{\mf{b}}^\kappa\}_{\kappa=1}^{n_r}$
to control the video context according to their preference. Here, $\widehat{\mf{b}}^\kappa \in\mathbb{R}^{n_v\times 4}$ is the bounding box, which is used to generate personalized concept associated with the $\kappa$-th region text prompt $\widehat{\mf{p}}^\kappa$. $n_v$ is the number of frames in the synthesized videos. After obtaining the layer-specific textual embedding $\widehat{\mathbf{c}}^\kappa = \{\widehat{\mathbf{c}}_l^\kappa \in\mathbb{R}^{n_e\times d}\}_{l=1}^L$ for $\widehat{\mf{p}}^\kappa$ using $\Phi(\cdot)$, we leverage $\{\widehat{\mathbf{c}}_l\}_{l=1}^L$ and the new 3D UNet $\epsilon_{\theta_*}(\cdot)$ to extract the latent features $\{\mf{f}_l \in\mathbb{R}^{n_v\times h_l\times w_l\times d}\}_{l=1}^{L}$ for $\widehat{\mf{p}}$. $w_l$ and $h_l$ are the width and height of the feature $\mf{f}_l$. At the $l$-th layer, we further conduct layer-specific region attention between $\mf{f}_l$ and $\widehat{\mathbf{c}}_l^\kappa$ to derive the $\kappa$-th region feature $\mf{f}_l^\kappa
\in\mathbb{R}^{n_v\times h_l^\kappa\times w_l^\kappa\times d}$:
\begin{align}
\mf{f}_l^\kappa = \mathcal{A}_l^\kappa \cdot \mathbf{V}_l^\kappa= \sigma(\mathbf{Q}_l(\mathbf{K}_l^\kappa)^\top/\sqrt{d} \cdot \mathbf{V}_l^\kappa,
\label{eq: feature_l_layer}
\end{align}
where $\mathcal{A}_l^\kappa = \sigma(\mathbf{Q}_l(\mathbf{K}_l^\kappa)^\top/\sqrt{d}\in\mathbb{R}^{n_v\times h_l^\kappa\times w_l^\kappa\times n_e}$ denotes the $\kappa$-th region attention, $\sigma(\cdot)$ is the sigmoid function, $w_l^\kappa$ and $h_l^\kappa$ represent the width and height of the $\kappa$-th bounding box $\widehat{\mf{b}}^\kappa$. Moreover, $\mathbf{Q}_l \in \mathbb{R}^{n_v\times h_l^\kappa\times w_l^\kappa\times d}, \mathbf{K}_l^\kappa \in\mathbb{R}^{n_e\times d}$ and $\mathbf{V}_l^\kappa \in\mathbb{R}^{n_e\times d}$ denote the query, key and value matrices:
\begin{align}
\mathbf{Q}_l=\Gamma(\mathbf{f}_l \mathbf{w}_q \otimes \widehat{\mathbf{m}}_l^\kappa), \mathbf{K}_l^\kappa = \widehat{\mathbf{c}}_l^\kappa \mathbf{w}_k, \mathbf{V}_l^\kappa = \widehat{\mathbf{c}}_l^\kappa \mathbf{w}_v,
\label{eq: QKV}
\end{align}
where $\widehat{\mathbf{m}}_l^\kappa\in\mathbb{R}^{n_v\times h_l\times w_l}$ represents the binary mask of the $\kappa$-th region, with all values within the $\kappa$-th bounding box $\widehat{\mf{b}}^\kappa$ assigned to $1$. $\otimes$ denotes the Hardmard product, $\Gamma(\cdot)$ preserves exclusively the features located within $\widehat{\mf{b}}^\kappa$. $\mathbf{w}_q, \mathbf{w}_k, \mathbf{w}_v\in \mathbb{R}^{d\times d}$ are the projection matrices in $\epsilon_{\theta_*}(\cdot)$. 

Subsequently, we substitute the values of $\mf{f}_l$ within the bounding boxes $\{\widehat{\mf{b}}^\kappa\}_{\kappa=1}^{n_r}$ with their corresponding region features $\{\mf{f}_l^\kappa\}_{\kappa=1}^{n_r}$, resulting in a new feature map $\widehat{\mathbf{f}}_l$ at the $l$-th layer. The layer-specific region attention is then applied to every layer of $\epsilon_{\theta_*}(\cdot)$ for multi-concept video customization.  

\begin{figure*}[t]
\centering
\includegraphics[width =1.0\linewidth]
{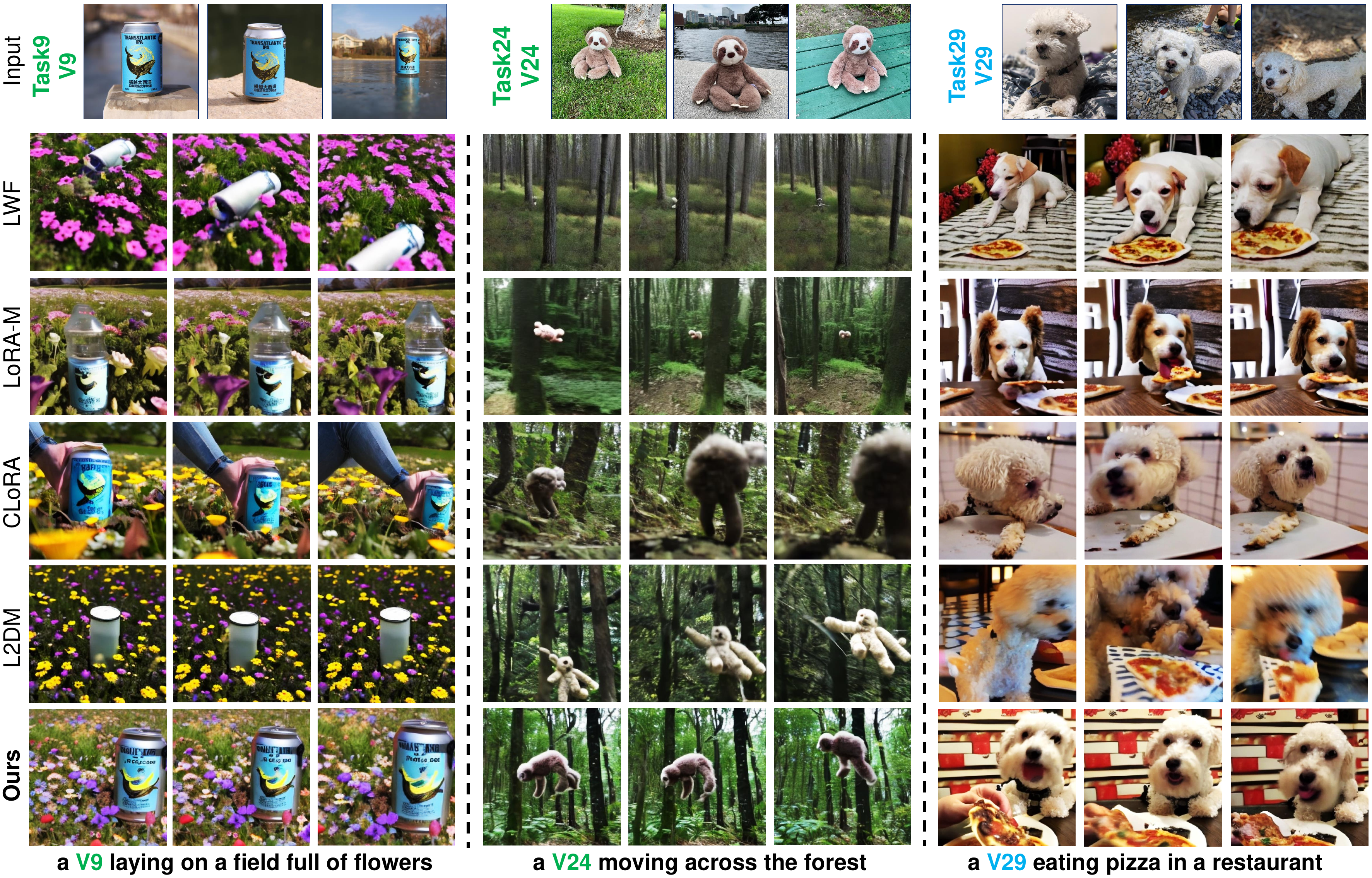}
\vspace{-7mm}
\caption{Comparison results of single-concept video customization under the CTVC setting when the backbone is DreamVideo. }
\label{fig: comparison_single_concept} 
\vspace{-3mm} 
\end{figure*}

\textbf{Attention-Guided Noise Estimation:} Given the initial text prompt $\widehat{\mf{p}}$, motivated by \cite{ho2021classifierfree}, we can derive the noise estimation $\mathbf{E}_t = \epsilon_{\theta_*} (\mathbf{z}_t, t) + \tau \cdot (\epsilon_{\theta_*} (\mathbf{z}_t, \widehat{\mathbf{c}}, t) - \epsilon_{\theta_*} (\mathbf{z}_t, t)) \in\mathbb{R}^{n_v\times h_L\times w_L\times d_L}$ at the $t$-th timescale, by incorporating the aforementioned layer-wise region attention into $\epsilon_{\theta_*}(\cdot)$. Here, $\mf{E}_t$ is the output of the $L$-th layer of $\epsilon_{\theta_*}(\cdot)$, and $\tau=7.5$ is the scaling factor. $w_L, h_L$ and $d_L$ represent the width, height and channel dimension of the noise estimation $\mathbf{E}_t$. $\epsilon_{\theta_*} (\mathbf{z}_t,t)$ denotes the unconditional noise estimation, while $\epsilon_{\theta_*} (\mathbf{z}_t, \widehat{\mathbf{c}}, t)$ represent noise estimation conditioned on $\widehat{\mathbf{c}}$. As a result, the $\kappa$-th ($\kappa=1, \cdots, n_r$) region noise estimation $\mathbf{E}_t^\kappa \in\mathbb{R}^{n_v\times h_L\times w_L\times d_L}$ is expressed:
\begin{align}
\!\!\mathbf{E}_t^\kappa \!=\! \epsilon_{\theta_*} (\mathbf{z}_t,t) + \tau \cdot \big(\epsilon_{\theta_*} (\mathbf{z}_t, [\widehat{\mathbf{c}}^\kappa, \widehat{\mathbf{b}}^\kappa], t) - \epsilon_{\theta_*} (\mathbf{z}_t,t) \big),\!\!
\label{eq: region_noise_estimation}
\end{align}
where $\epsilon_{\theta_*} (\mathbf{z}_t, [\widehat{\mathbf{c}}^\kappa, \widehat{\mathbf{b}}^\kappa], t)$ represents the noise estimation based on the $\kappa$-th regional condition $[\widehat{\mathbf{c}}^\kappa, \widehat{\mathbf{b}}^\kappa]$.

To address the challenge of concept neglect in multi-concept video customization, we aim to place greater emphasis on personalized concepts that are easily overlooked. In light of this, we propose an attention-guided noise estimation module that leverages region attention to assign larger weights to easily overlooked concepts when aggregating $n_r$ region estimations $\{\mf{E}_t^\kappa\}_{\kappa=1}^{n_r}$. Specifically, for the $\kappa$-th ($\kappa=1, \cdots, n_r$) region condition, we upsample $L$ layer-specific region attentions $\{\mathcal{A}_l^\kappa \in \mathbb{R}^{n_v\times h_l^\kappa\times w_l^\kappa\times n_e}\}_{l=1}^L$ to $\{\widehat{\mathcal{A}}_l^\kappa \in\mathbb{R}^{n_v\times h_L^\kappa\times w_L^\kappa\times n_e}\}_{l=1}^L$, and average them to obtain $\widehat{\mathcal{A}}^\kappa = \frac{1}{L}\sum_{l=1}^L \widehat{\mathcal{A}}_l^\kappa$. Subsequently, we select the maximum attention value $\omega^\kappa = \max(\widehat{\mathcal{A}}^\kappa) \in\mathbb{R}^{n_v}$ to reweight the noise estimation $\mf{E}_t^{\kappa}$. Intuitively, if the attention value $\omega^\kappa$ of the $\kappa$-th region ($\kappa=1, \dots, n_r$) is small, it indicates that the corresponding concept is more likely to be overlooked in video customization. Thus, we assign more weight to the noise estimation with a small attention value. To achieve this, we concatenate $\{\omega^\kappa\}_{\kappa=1}^{n_r}$ as $\omega \in\mathbb{R}^{n_v\times n_r}$ and normalize $\omega$ as $\widehat{\omega}\in\mathbb{R}^{n_v\times n_r}$. To resolve concept neglect, we employ $\widehat{\omega}$ to aggregate $n_r$ region noise estimation to obtain $\widehat{\mf{E}}_t$:
\begin{align}
\widehat{\mathbf{E}}_t = \mathbf{E}_t + \sum\nolimits_{\kappa=1}^{n_r} \big(\mathbf{1} - \mathrm{EP}(\widehat{\omega}[\kappa]) \big)\otimes \mathbf{E}_t^\kappa \otimes \widehat{\mathbf{m}}_L^\kappa,
\label{eq: noise_estimation_aggregation} 
\end{align}
where $\widehat{\omega}[\kappa] \in\mathbb{R}^{n_v}$ is the $\kappa$-th column of $\widehat{\omega}$, and $\mathrm{EP}(\widehat{\omega}[\kappa])$ expands $\widehat{\omega}[\kappa]$ as $\mathbb{R}^{n_v\times h_L\times w_L\times d_L}$. $\widehat{\mathbf{m}}_L^\kappa \in\mathbb{R}^{n_v\times h_L\times w_L}$ represents the binary region mask corresponding to the bounding box $\widehat{\mf{b}}^\kappa \in\mathbb{R}^{n_v\times 4}$ at the $L$-th layer of $\epsilon_{\theta_*}(\cdot)$. Finally, we follow \cite{10656166} and feed $\widehat{\mathbf{E}}_t$ into UNet $\epsilon_{\theta_*}(\cdot)$ to perform multi-concept customization via gradual denoising. 

\begin{figure*}[t]
\centering
\includegraphics[width =1.0\linewidth]
{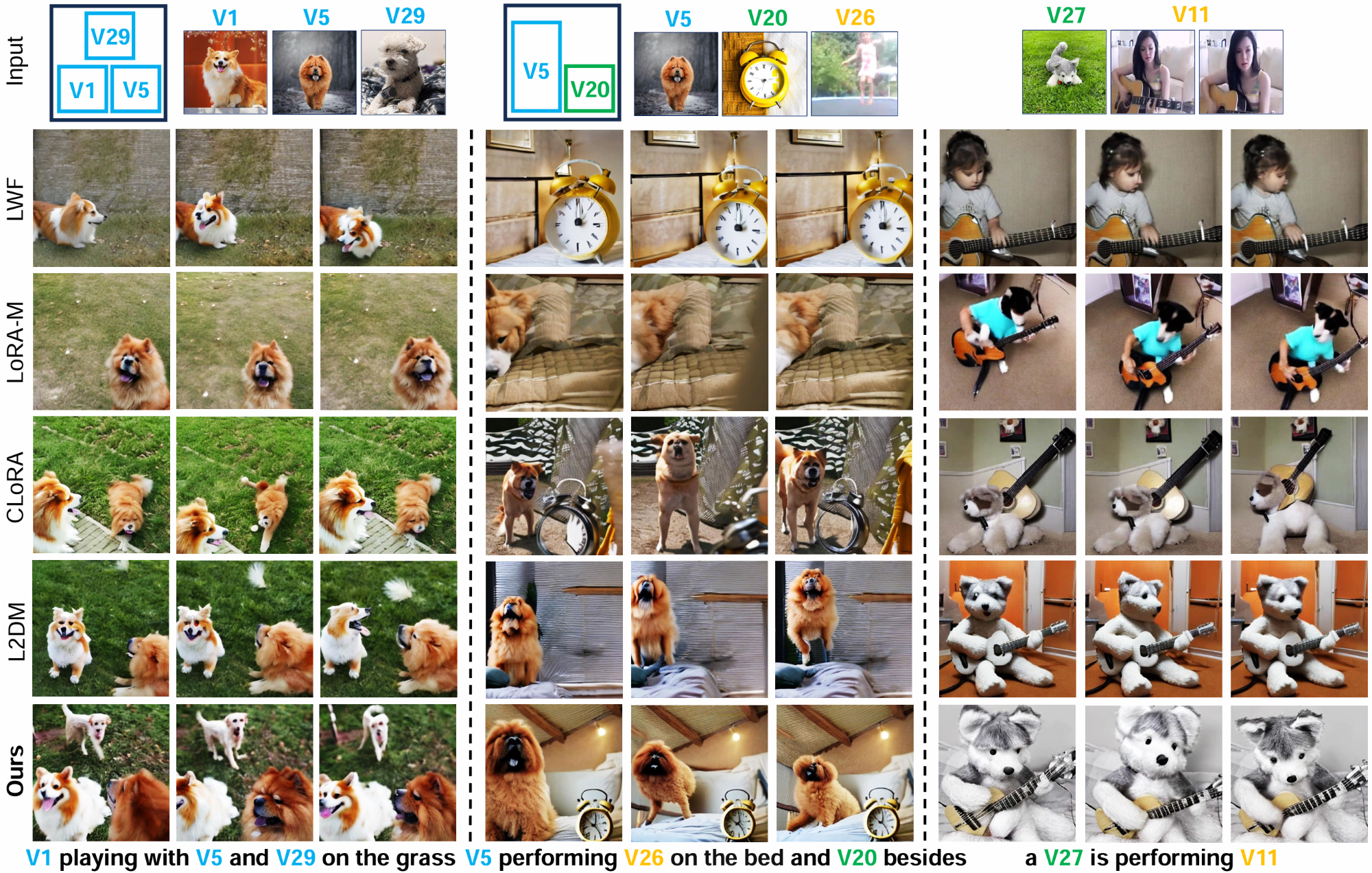}
\vspace{-7mm}
\caption{Comparison results of multi-concept video customization under the CTVC setting when the backbone is DreamVideo. }
\label{fig: compar_multi_concept}
\vspace{-3mm}
\end{figure*}

\begin{figure}[t]
\centering
\includegraphics[width = 1.0\linewidth]
{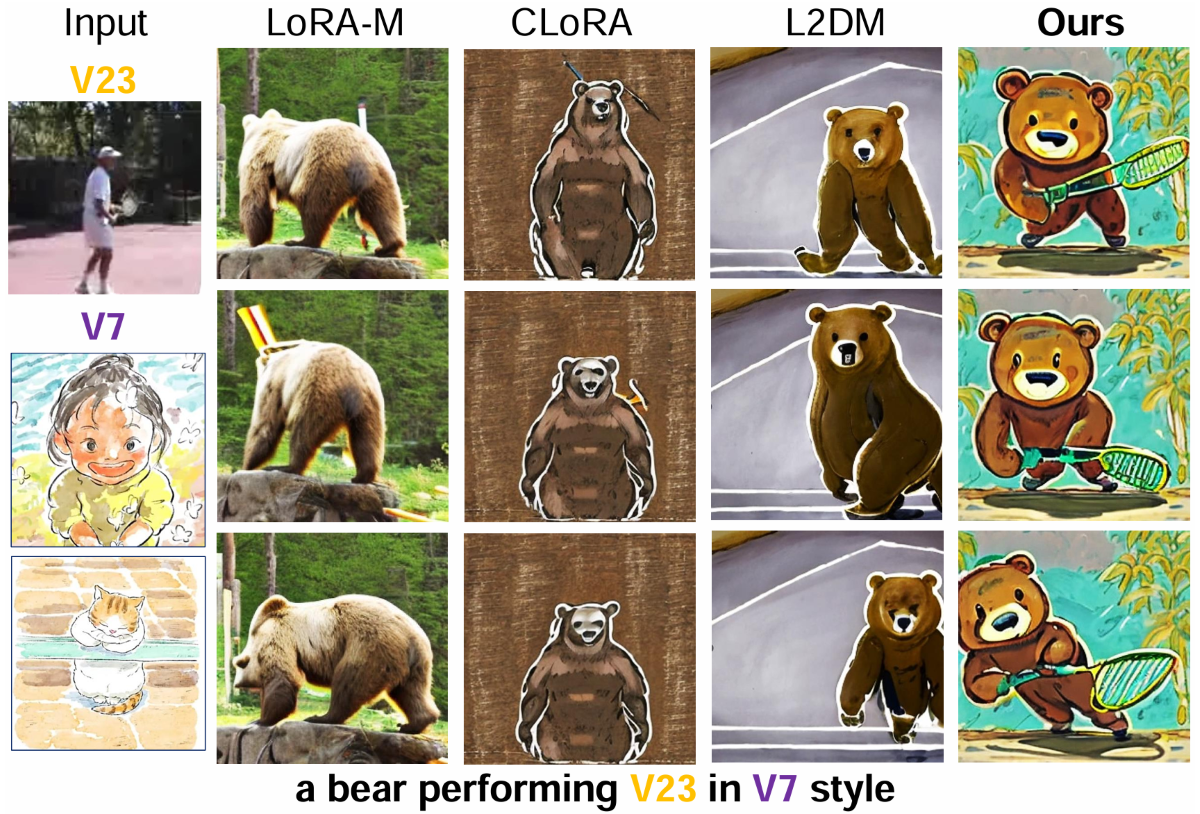}
\vspace{-7mm}
\caption{Comparisons of style transfer under CTVC setting when the backbone is DreamVideo. }
\label{fig: compar_style_transfer}
\end{figure}

\section{Experiments}
\subsection{Implementation Details}
\textbf{Dataset:}
Inspired by \cite{1011453687945, 10656166}, we introduce a new continual video customization (CVC) dataset to evaluate the effectiveness of our model under the CTVC setting. 
In the CVC dataset, there are 35 sequential text-to-video customization tasks, including 25 personalized subjects and 10 motion patterns. Among the 25 personalized subjects, 9 are pets, 5 are styles, and the remaining 11 are objects. To meet the practical requirements of CTVC, during training, we set one text-video pair for motions tasks and $3\sim5$ image-text pairs for subjects tasks. More importantly, we include several semantically similar concepts (\emph{e.g.}, seven dog and two cat) to make the dataset more challenging.

\begin{figure}[t]
\centering
\includegraphics[width =1.0\linewidth]
{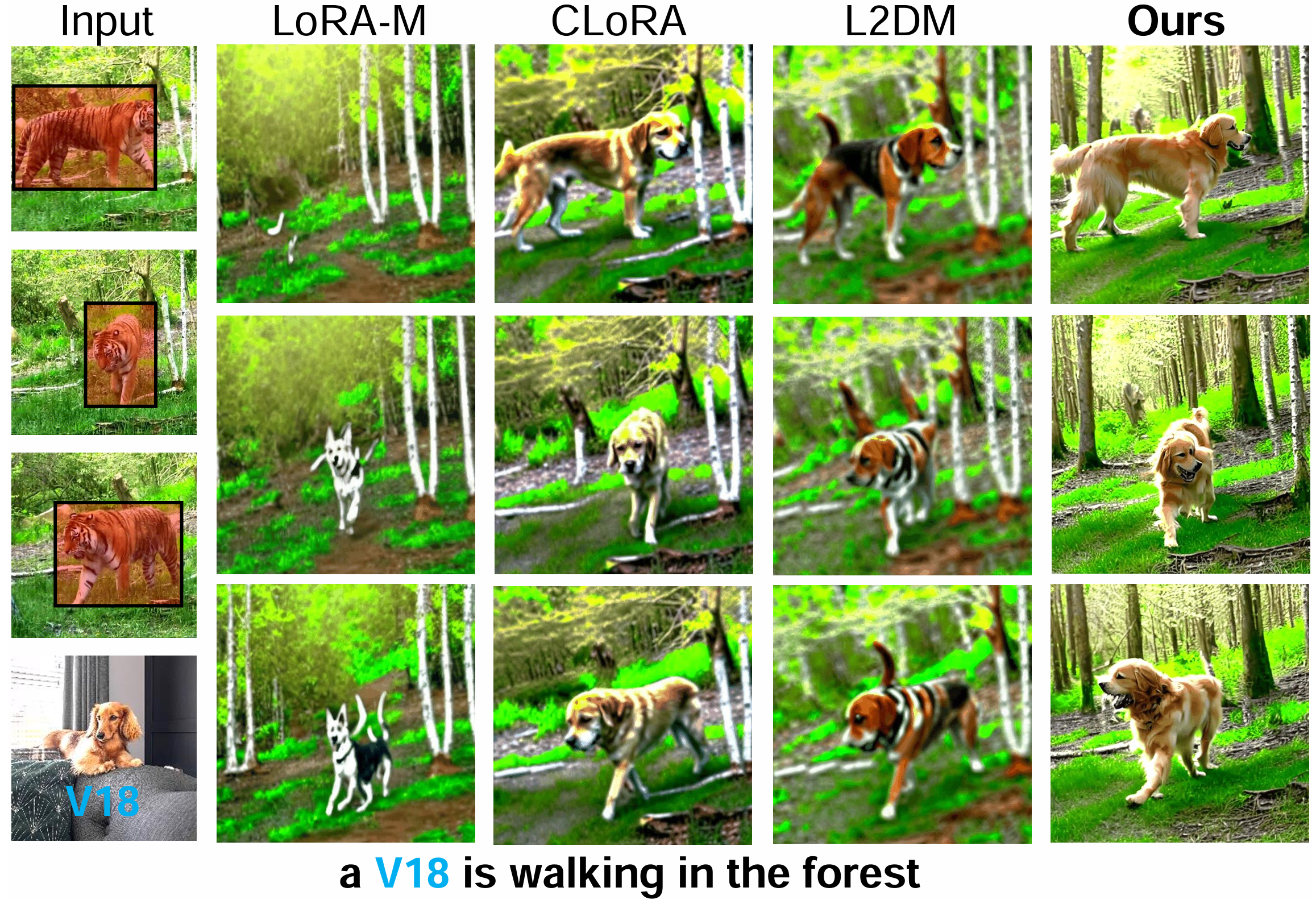}
\vspace{-7mm}
\caption{Comparisons of video editing under CTVC setting when the backbone is DreamVideo. }
\label{fig: compar_video_editing}
\end{figure}

\begin{figure*}[t]
\centering
\includegraphics[width =1.0\linewidth]
{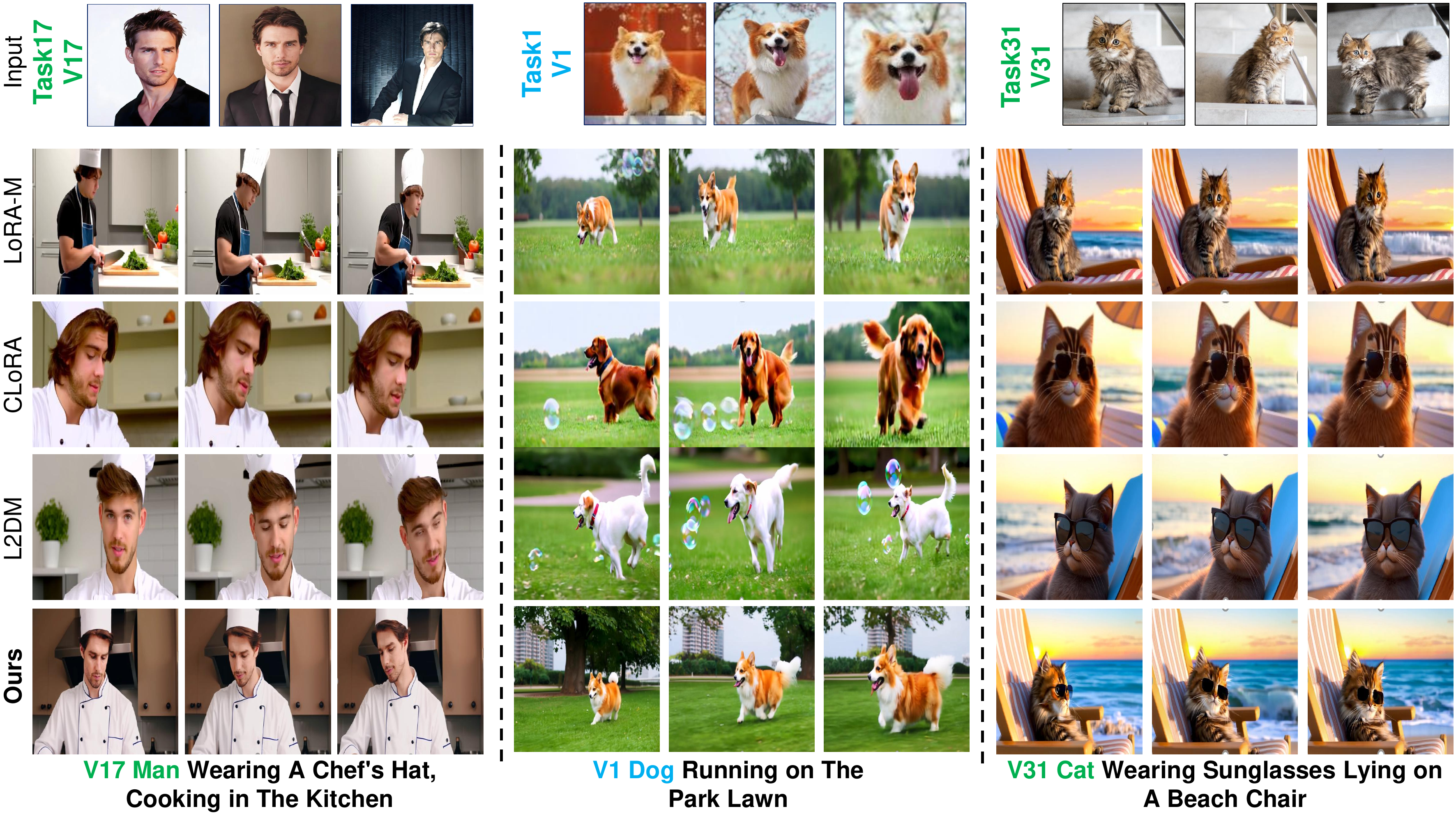}
\vspace{-7mm}
\caption{Comparison results of single-concept video customization under the CTVC setting when the backbone is Wan 2.1. }
\label{fig: compar_single_concept_wan}
\end{figure*}

\begin{figure*}[t]
\centering
\includegraphics[width =1.0\linewidth]
{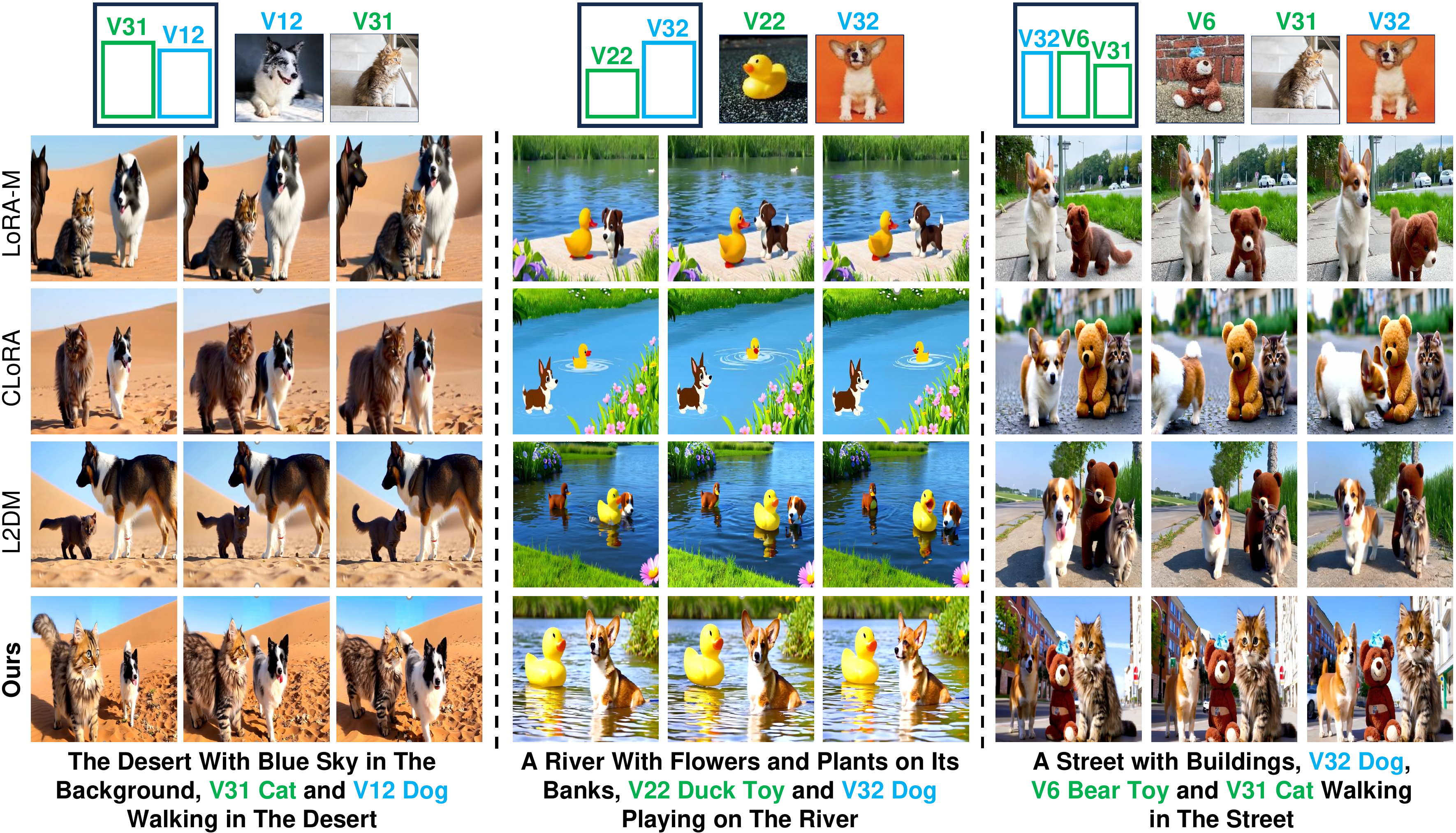}
\vspace{-7mm}
\caption{Comparison results of multi-concept video customization under the CTVC setting when the backbone is Wan 2.1. }
\label{fig: compar_multi_concept_wan}
\end{figure*}

\begin{table}[t]
\centering
\setlength{\tabcolsep}{0.5pt}
\renewcommand{\arraystretch}{0.8}
\caption{Results of single-concept video customization when the backbone is DreamVideo.}
\vspace{-3mm}
\resizebox{\linewidth}{!}{
\begin{tabular}{l|ccccccc}
\toprule
\makecell[c]{Methods} & ~CLIPT~ & ~CLIPI~ & ~DINOI~ & ~TCons~ & ~F-CLIPT~& ~F-CLIPI~ & ~F-DINO~ \\
\midrule
Finetuning & 22.9 & 52.3 & 25.1 & 96.3 & 9.5 & 14.5 & 11.2 \\

EWC & 28.1 & 58.4 & 29.5 & 96.1 & 5.4 &\ \ 7.8 & \ \ 6.3 \\

LWF & 24.8 & 59.3 & 30.0 & 96.2 & 4.3 & \ \ 6.8 & \ \ 5.8 \\

LoRA-M  & 25.5 & 57.8 & 29.0 & 96.1 & 5.2 & \ \ 8.4 & \ \ 6.2 \\

CLoRA & 28.7 & 60.7 & 31.0 & 96.1 & 3.5 & \ \ 5.3 & \ \ 4.1 \\

L2DM  & 28.3 & 62.3 & 31.2 & 96.2 & 3.1 & \ \ 3.9 & \ \ 3.7 \\
\midrule
\textbf{CCVD}  & \textbf{29.9} & \textbf{64.3} & \textbf{33.4} & \textbf{96.4} & \textbf{1.1} & \ \ \textbf{2.3} & \ \ \textbf{1.9} \\
\bottomrule
\end{tabular}}
\label{tab: quantitative_results}
\vspace{-3mm}
\end{table}

\begin{table}[t!]
\centering
\setlength{\tabcolsep}{1.3pt}
\renewcommand{\arraystretch}{0.8}
\caption{Results (CLIPI) of single-concept customization when the backbone is DreamVideo. }
\vspace{-3mm}
\resizebox{\linewidth}{!}{
\begin{tabular}{l|cccccccc}
\toprule
\makecell[c]{Methods} & V1-5 & V6-10 & V11-15 & V15-20 & V21-25 & V25-30 & V31-35 & ~Avg.~ \\
\midrule
Finetuning & 57.0 & 50.2 & 51.8 & 52.4 & 50.9 & 51.4 & 52.0 & 52.3 \\

EWC & 63.5 & 57.7 & 59.2 & 58.4 & 56.6 & 55.8 & 57.6 & 58.4 \\

LWF & 64.4 & 58.2 & 62.2 & 59.2 & 56.4 & 56.4 & 58.5 & 59.3 \\

LoRA-M & 62.6 & 58.1 & 58.1 & 57.9 & 55.2 & 55.5 & 57.2 & 57.8 \\

CLoRA & 66.1 & 59.9 & 63.2 & 60.5 & 57.8 & 57.7 & 59.3 & 60.7\\

L2DM & 67.1 & 60.1 & 64.2 & 63.6 & 62.7 & 61.1 & 61.8 & 62.3 \\

\midrule
\textbf{CCVD}  & \textbf{68.7} & \textbf{61.6} & \textbf{66.4} & \textbf{64.9} & \textbf{63.6} & \textbf{62.0} & \textbf{62.5} & \textbf{64.3}\\
\bottomrule
\end{tabular}}
\label{tab: quantitative_compare_task_wise_CLIPI}
\end{table}

\textbf{Setups:}
For fair comparisons, our model and all comparative methods utilize DreamVideo \cite{10656166} and Wan 2.1 \cite{wan2025wanopenadva} as the baseline backbones. For DreamVideo \cite{10656166}, we use the Adam optimizer for training, with an initial learning rate of $1.0\times 10^{-4}$ for textual embeddings and $1.0\times 10^{-5}$ for the 3D UNet. For Wan 2.1 \cite{wan2025wanopenadva}, the initial learning rate for both the textual embeddings and the diffusion transformer backbone is set to $1.0 \times 10^{-4}$. Besides, we set $\lambda=0.1$ in Eq.~\eqref{eq: COL_loss}, and $n_v=32$ to synthesize customized videos. Besides, we perform comprehensive qualitative and quantitative evaluations across various generation tasks including single/multi-concept video customization, style transfer and video editing. For quantitative comparisons, we follow \cite{10656166} to utilize CLIPT, CLIPI, DINOI, TCons as evaluation metrics. Moreover, we use the forgetting of CLIPT, CLIPI, and DINOI (denoted as F-CLIPT, F-CLIPI, and F-DINOI) \cite{sun2024create} to assess CTVC performance.

\subsection{Qualitative Comparisons} 
To demonstrate the superiority of our CCVD model, as in Fig.~\ref{fig: comparison_single_concept}--\ref{fig: compar_multi_concept_wan}, we conduct qualitative comparisons on single/multi-concept video customization, style transfer and video editing. 1) In Figs.~\ref{fig: comparison_single_concept} and \ref{fig: compar_single_concept_wan}, our model surpasses other methods in single-concept video customization, since it can mitigate forgetting by incorporating concept-specific attribute retention module to explore unique concept identity. 2) In Fig.~\ref{fig: compar_multi_concept} and \ref{fig: compar_multi_concept_wan}, all comparative methods fail to generate multiple concepts, while our model can leverage layer-specific region attention and attention-guided noise estimation to synthesize videos with multiple user-specified concepts. 3) CCVD achieves superior performance in video style transfer (Fig.~\ref{fig: compar_style_transfer}), as our task-aware concept aggregation module can better maintain style integrity. 4) In Fig.~\ref{fig: compar_video_editing}, we integrate Video-P2P \cite{10656631} as a plugin for editing. The large improvement of CCVD compared to other methods is attributed to the concept-specific attribute retention module, which can preserve unique concept.

\subsection{Quantitative Comparisons} 
To conduct quantitative comparisons between our model and other methods, we follow the evaluation protocol proposed in \cite{1011453687945}. Specifically, we set 10 prompts per concept and generate 10 videos per prompt, resulting in a total of 100 synthesized videos for performance assessment. As shown in Tabs.~\ref{tab: quantitative_results}--\ref{tab: quantitative_compare_task_wise_CLIPI}, our proposed model surpasses competing methods across all evaluation metrics, which verifies the superiority of our model to resolve the CTVC problem. This indicates that our model effectively preserves the unique identity of each learned concept via the concept-specific attribute retention module. Moreover, proposed task-aware concept aggregation can dynamically aggregate the subject and motion adapters to alleviate forgetting during the testing phase. 

\begin{table}[t!]
\centering
\setlength{\tabcolsep}{1.3pt}
\renewcommand{\arraystretch}{1}
\caption{Ablation of single-concept video customization when the backbone is DreamVideo. }
\vspace{-3mm}
\resizebox{\linewidth}{!}{
\begin{tabular}{l|ccc|ccccc}
\toprule
\makecell[c]{Methods} & TCA & LCT & COL & ~CLIPT~ & ~CLIPI~ & ~DINOI~ & ~F-CLIPT~ & ~F-CLIPI~ \\
\midrule
Baseline & \xmark & \xmark & \xmark & 25.5 & 57.8 & 29.0 & 5.2 & 8.4 \\
Baseline w/ TCA & \cmark & \xmark & \xmark & 28.7 & 62.2 & 31.3 & 3.4 & 4.1 \\
Ours w/o COL & \cmark & \cmark & \xmark  & 29.8 & 63.6 & 33.0 & 1.3 & 2.4 \\
\midrule
\textbf{Ours} (\textbf{CCVD}) & \cmark & \cmark & \cmark & \textbf{29.9} & \textbf{64.3} & \textbf{33.4} & \textbf{1.1} & \textbf{2.3} \\

\bottomrule
\end{tabular} }
\label{tab: ablation_single_concept}
\vspace{-1mm}
\end{table}

\subsection{Ablation Study}
\begin{figure}[t!]
\centering
\includegraphics[width =1.0\linewidth]
{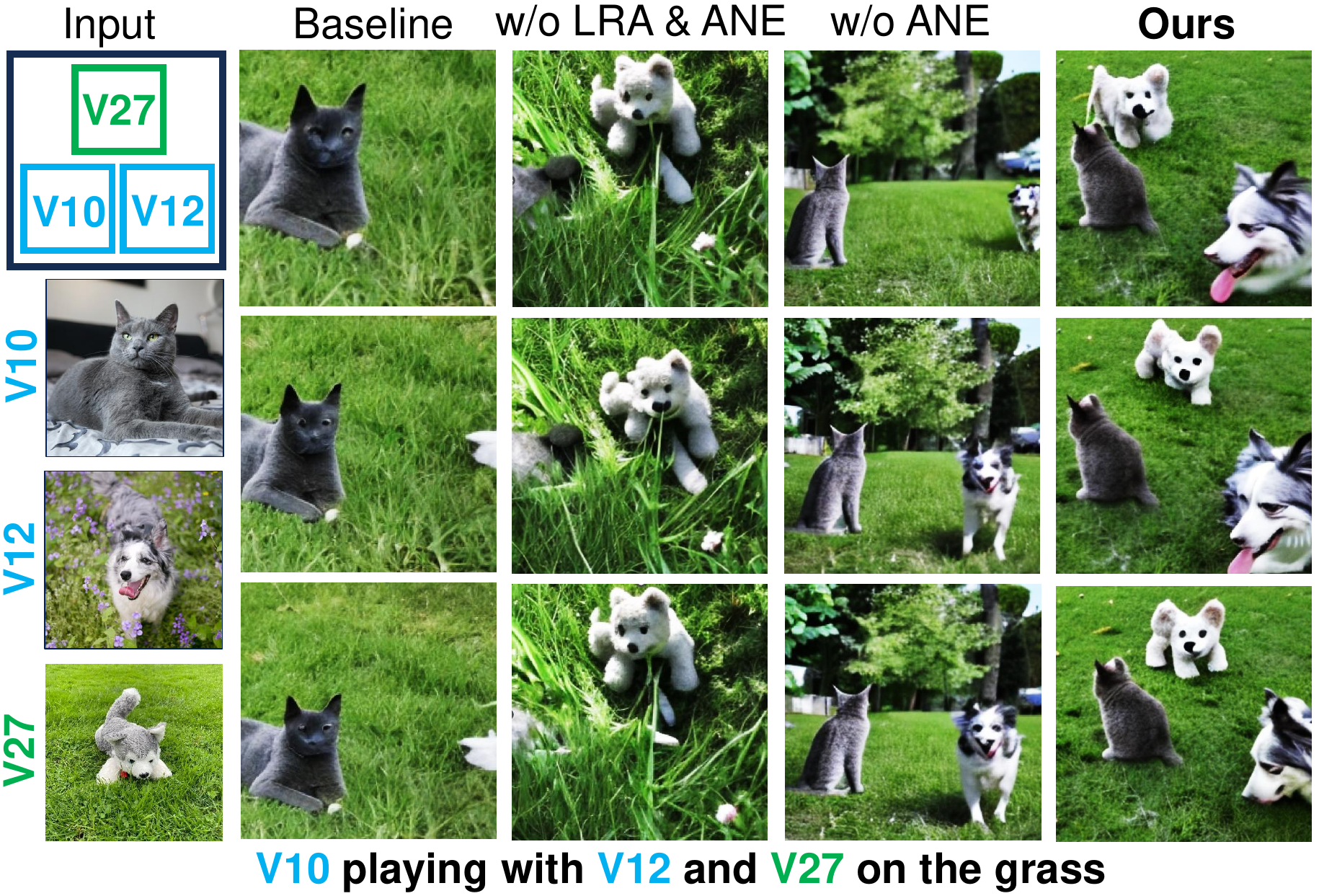}
\vspace{-7mm}
\caption{Ablation of multi-concept video customization when the backbone is DreamVideo. }
\label{fig: ablation_multi_concept}
\end{figure}

To verify the efficacy of all proposed modules, we conduct comprehensive ablation experiments in terms of layer-specific concept tokens (LCT), concept orthogonal loss (COL),
task-aware concept aggregation (TCA), layer-specific region attention (LRA) and attention-guided noise estimation (ANE). As shown in Tab.~\ref{tab: ablation_single_concept}, compared to ``Baseline'', our model achieves significant performance improvements across all evaluation metrics after incorporating the proposed LCT, COL, and TCA modules for single-concept video customization. It illustrates that our model effectively mitigates forgetting by capturing distinctive attributes of old concepts while dynamically aggregating subject and motion adapters. As presented in Fig.~\ref{fig: ablation_multi_concept}, the performance of our model in multi-concept video customization significantly declines when the LRA and ANE modules are removed. This confirms their effectiveness in mitigating the issue of concept neglect.

\section{Conclusion}
In this work, we propose a new problem termed Continual Text-to-Video Customization (CTVC), and develop a novel Continual Customized Video Diffusion (CCVD) model to resolve the CTVC problem. Specifically, we introduce a concept-specific attribute retention module during training and a task-aware concept aggregation during testing to alleviate catastrophic forgetting. Additionally, we design a controllable conditional synthesis module to address concept neglect by incorporating layer-specific region attention and attention-guided noise estimation. Comprehensive comparisons show that our CCVD has superiority performance.

\section*{Acknowledgments}
This work was supported by the National Key Research and Development Program of China under Grant 2024YFB4707700, and the National Natural Science Foundation of China under Grant U23A20343, and CAS Project for Young Scientists in Basic Research, Grant No. YSBR-041, and Liaoning Provincial ``Selecting the Best Candidates by Opening Competition Mechanism" Science and Technology Program under Grant 2023JH1/10400045.
  
\bibliography{aaai2026}

\appendix
\section{Supplementary Material}

\subsection{Implementation Details}
\begin{figure*}[t]
\centering
\includegraphics[width =1.0\linewidth]
{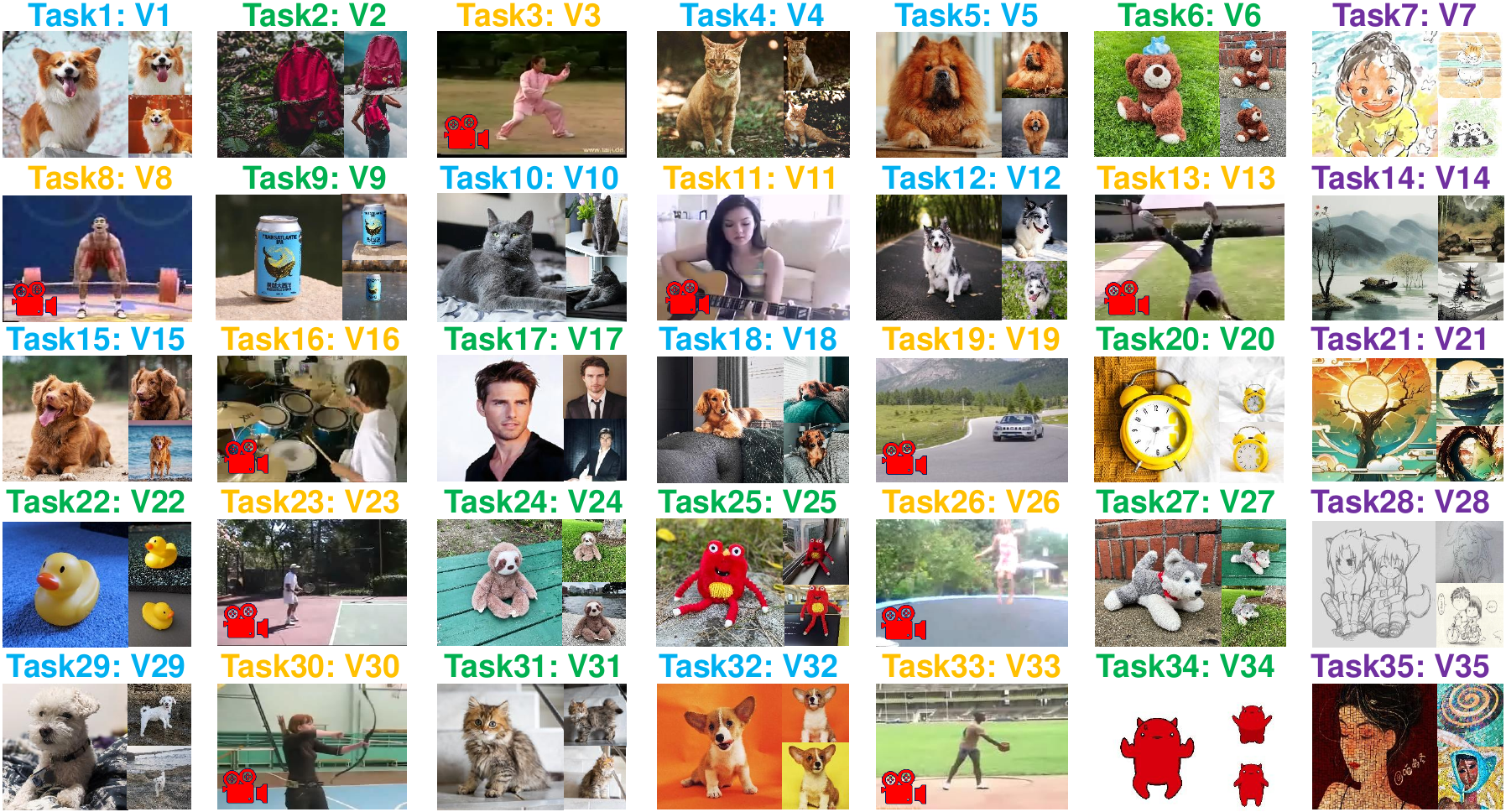}
\vspace{-7mm}
\caption{Visualization of 35 consecutive text-to-video customization tasks in the CVC dataset. }
\label{fig: vis_dataset}
\end{figure*}

\textbf{Dataset:}
Inspired by \cite{1011453687945, 10656166}, we introduce a new continual video customization (CVC) dataset to evaluate the effectiveness of our model under the CTVC setting. 
As shown in Fig.~\ref{fig: vis_dataset}, there are 35 sequential text-to-video customization tasks in the CVC dataset, including 25 personalized subjects and 10 motion patterns. Among the 25 personalized subjects, 9 are pets, 5 are styles, and the remaining 11 are objects. To satisfy the practical requirements of the CTVC setting, during training, we set one text-video pair for tasks involving motions and $3\sim5$ image-text pairs for tasks involving subjects. More importantly, we include several semantically similar personalized concepts (\emph{e.g.}, seven dog concepts and two cat concepts) to make the CVC dataset more challenging.

\begin{figure*}[t]
\centering
\includegraphics[width =1.0\linewidth]
{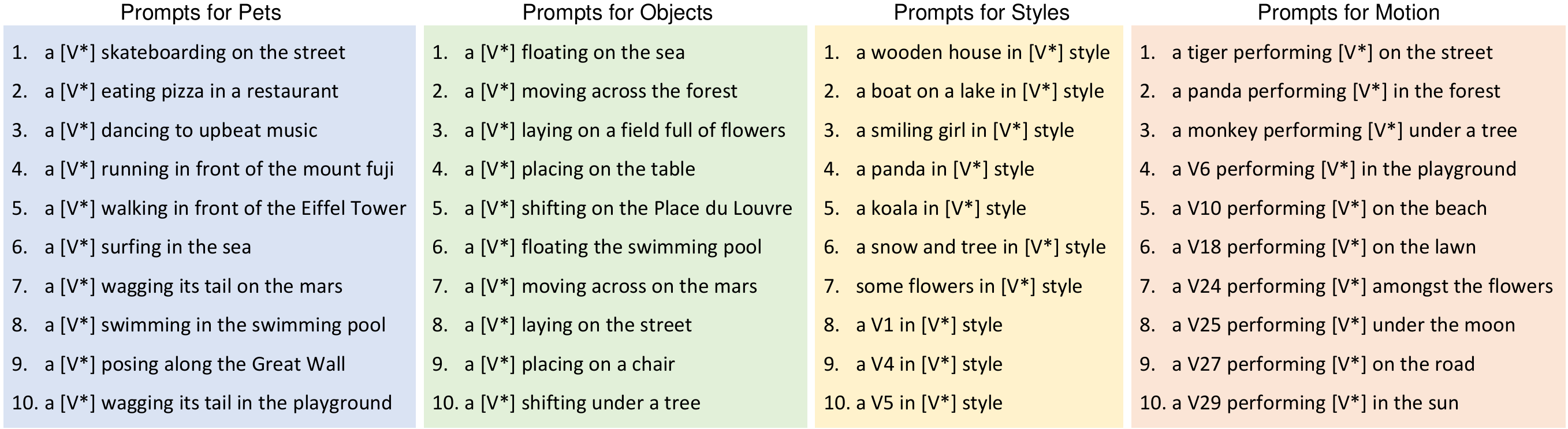}
\vspace{-7mm}
\caption{Text prompts used in the paper for quantitative comparisons. }
\label{fig: vis_prompts}
\end{figure*}

\textbf{Setups:}
For fair comparisons, our proposed model and all comparative methods utilize DreamVideo \cite{10656166} as the baseline backbone. For DreamVideo \cite{10656166}, we use the Adam optimizer for training, with an initial learning rate of $1.0\times 10^{-4}$ for textual embeddings and $1.0\times 10^{-5}$ for the 3D UNet. For Wan 2.1 \cite{wan2025wanopenadva}, the initial learning rate for both the textual embeddings and the diffusion transformer backbone is set to $1.0 \times 10^{-4}$. In this paper, we set $\lambda=0.1$ in Eq.~\eqref{eq: COL_loss}. Furthermore, we set the training steps to $1,000$ for subject and motion adapters and $3,000$ for concept tokens. 
During the inference phase, we follow DreamVideo \cite{10656166} to employ the DDIM sampler \cite{song2021denoising} with 50 steps and apply classifier-free guidance \cite{ho2021classifierfree} across all baseline models. Moreover, videos are generated with a spatial resolution of $256\times 256$, consisting of $32$ frames at a rate of 8 frames per second (fps). All comparison experiments are performed on a single NVIDIA 3090 GPU.

\begin{table}[b]
\centering
\setlength{\tabcolsep}{1.3pt}
\renewcommand{\arraystretch}{0.8}
\caption{Results (CLIPI) of single-concept customization when the backbone is DreamVideo. }
\vspace{-3mm}
\resizebox{1.0\linewidth}{!}{
\begin{tabular}{l|cccccccc}
\toprule
\makecell[c]{Methods} & V1-5 & V6-10 & V11-15 & V15-20 & V21-25 & V25-30 & V31-35 & ~Avg.~ \\
\midrule
Finetuning & 21.9 & 23.1 & 29.8 & 22.9 & 23.0 & 22.6 & 23.0 & 22.9 \\

EWC & 30.0 & 28.1 & 28.8 & 28.5 & 27.6 & 27.6 & 25.8 & 28.1 \\

LWF& 29.3 & 24.1 & 23.7 & 25.7 & 24.2 & 23.1 & 23.7 & 24.8 \\

LoRA-M  & 27.6 & 25.5 & 25.2 & 25.7 & 25.9 & 23.7 & 24.6 & 25.5 \\

CLoRA  & 30.3 & 29.0 & 29.4 & 28.8 & 27.9 & 28.3 & 27.0 & 28.7 \\

L2DM  & 29.6 & 28.2 & 29.2 & 28.1 & 27.7 & 27.9 & 26.9 & 28.3 \\
\midrule
\textbf{CCVD}  & \textbf{31.1} & \textbf{29.7} & \textbf{30.6} & \textbf{30.2} & \textbf{28.9} & \textbf{29.0} & \textbf{29.3} & \textbf{29.9} \\
\bottomrule
\end{tabular}}
\label{tab: quantitative_compare_task_wise_CLIPT}
\end{table}

\textbf{Evaluation Metrics:}
We perform comprehensive qualitative and quantitative evaluations across various text-to-video generation tasks including single/multi-concept video customization, style transfer and video editing. 
As illustrated in Fig.~\ref{fig: vis_prompts}, we broadly classify prompts into four groups based on the concepts they encompass. To conduct quantitative comparisons between our model and state-of-the-art (SOTA) methods, following the approach of \cite{1011453687945}, we utilize an evaluation prompt and a consistent negative prompt to generate 10 videos. Consequently, with 10 evaluation prompts provided in this study, we produce $100$ videos per concept to assess model performance. Motivated by \cite{10656166}, we utilize CLIPT, CLIPI, DINOI, TCons as evaluation metrics for quantitative comparisons. Moreover, we introduce the forgetting of CLIPT, CLIPI, and DINOI (denoted as F-CLIPT, F-CLIPI, and F-DINOI) to assess our model's performance in addressing catastrophic forgetting of previously learned concepts.

\textbf{Comparative Methods:}
To thoroughly showcase the superior performance of our model, we introduce six state-of-the-art (SOTA) comparison methods. They include three continual learning-based approaches (\emph{e.g.}, Finetuning, EWC \cite{kirkpatrick2017overcoming} and LWF \cite{li2017learning}), two continual diffusion models (\emph{e.g.}, CLoRA \cite{smith2023continual} and L2DM \cite{sun2024create}), and one multi-concept composition technology (LoRA-M \cite{zhong2024multi}).
Specifically, Finetuning focuses on optimizing subject and motion adapters to sequentially learn multiple personalized concepts. To mitigate catastrophic forgetting, EWC \cite{kirkpatrick2017overcoming} employs an elastic regularizer to constrain changes in network parameters, and LWF \cite{li2017learning} retains old training data to facilitate knowledge distillation for the current task. CLoRA \cite{smith2023continual} introduces a self-regularized low-rank adaptation mechanism to continuously learn new personalized concepts. L2DM \cite{sun2024create} constructs a long-term memory bank to reconstruct past personalized concepts and utilizes knowledge distillation to reduce catastrophic forgetting. LoRA-M \cite{zhong2024multi} combines all subject and motion adapters equally to retrain the latent diffusion model. More importantly, all comparison methods and our proposed model employ DreamVideo \cite{10656166} as the baseline backbone for fair comparisons.

\begin{table}[t]
\centering
\setlength{\tabcolsep}{1.3pt}
\renewcommand{\arraystretch}{1.2}
\caption{Results (DINOI) of single-concept customization when the backbone is DreamVideo. }
\vspace{-3mm}
\resizebox{\linewidth}{!}{
\begin{tabular}{l|cccccccc}
\toprule
\makecell[c]{Methods} & V1-5 & V6-10 & V11-15 & V15-20 & V21-25 & V25-30 & V31-35 & ~Avg.~ \\
\midrule
Finetuning & 28.1 & 22.4 & 31.5 & 27.3 & 22.5 & 24.7 & 19.3 & 25.1\\

EWC & 33.2 & 30.5 & 35.2 & 32.0 & 24.9 & 25.9 & 24.6 & 29.5\\

LWF& 33.9 & 31.9 & 36.2 & 32.8 & 25.5 & 25.6 & 24.6 & 30.0\\

LoRA-M & 32.9 & 30.2 & 35.0 & 31.9 & 24.7 & 24.3 & 24.0 & 29.0 \\

CLoRA & 34.3 & 33.1 & 37.6 & 32.6 & 27.2 & 26.5 & 25.4 & 31.0\\

L2DM  & 36.2 & 33.6 & 35.6 & 32.5 & 27.7 & 26.6 & 25.9  & 31.2 \\

\midrule
\textbf{CCVD}  & \textbf{38.3}& \textbf{34.8} & \textbf{39.1} & \textbf{35.5} & \textbf{28.4} & \textbf{29.7} & \textbf{28.0} & \textbf{33.4} \\

\midrule
\end{tabular}}
\label{tab: quantitative_compare_task_wise_DINOI}
\end{table}

\begin{table}[t]
\centering
\setlength{\tabcolsep}{1.3pt}
\renewcommand{\arraystretch}{1.2}
\caption{Results (F-CLIPI) of single-concept customization when the backbone is DreamVideo. }
\vspace{-3mm}
\resizebox{\linewidth}{!}{
\begin{tabular}{l|cccccccc}
\toprule
\makecell[c]{Methods} & V1-5 & V6-10 & V11-15 & V15-20 & V21-25 & V25-30 & V31-35 & ~Avg.~ \\
\midrule
Finetuning & 12.7 & 12.9 & 14.5 & 14.9 & 15.4 & 15.6 & 15.5 & 14.5 \\

EWC & 6.2 & 5.9 & 7.4 & 8.1 & 8.8 & 9.3 & 9.4 & 7.8 \\

LWF & 5.3 & 5.2 & 5.9 & 6.8 & 7.8 & 8.4 & 8.4 & 6.8 \\

LoRA-M  & 7.1 & 6.1 & 7.9 & 8.6 & 9.5 & 9.9 & 10.0 & 8.4 \\

CLoRA & 3.6 & 3.5 & 4.4 & 5.3 & 6.4 & 7.0 & 7.1 & 5.3 \\

L2DM & 2.6 & 2.9 & 3.7 & 4.0 & 4.3 & 4.7 & 4.8 & 3.9 \\
\midrule
\textbf{CCVD} & \textbf{1.0} & \textbf{1.3} & \textbf{1.9} & \textbf{2.4} & \textbf{2.8} & \textbf{3.3} & \textbf{3.5} & \textbf{2.3} \\
\midrule
\end{tabular}}
\label{tab: quantitative_compare_task_wise_FCLIPI}
\end{table}

\subsection{Quantitative Comparisons}
To perform quantitative evaluations between our proposed CCVD model and other competing approaches, we adhere to the evaluation framework outlined in \cite{1011453687945}. Specifically, we define 10 prompts for each concept and produce 10 videos per prompt, yielding a total of 100 generated videos for performance analysis. As demonstrated in Tabs.~\ref{tab: quantitative_compare_task_wise_CLIPT}--\ref{tab: quantitative_compare_task_wise_FCLIPI}, our proposed model outperforms competing methods across all evaluation metrics, confirming its superiority in addressing the CTVC problem. This highlights the effectiveness of our model in preserving the distinct identity of each learned concept through the concept-specific attribute retention module. Additionally, the proposed task-aware concept aggregation strategy dynamically combines the subject and motion adapters, effectively mitigating catastrophic forgetting of previously learned concepts during the testing phase.

\begin{figure*}
\centering
\includegraphics[width =1.0\linewidth]
{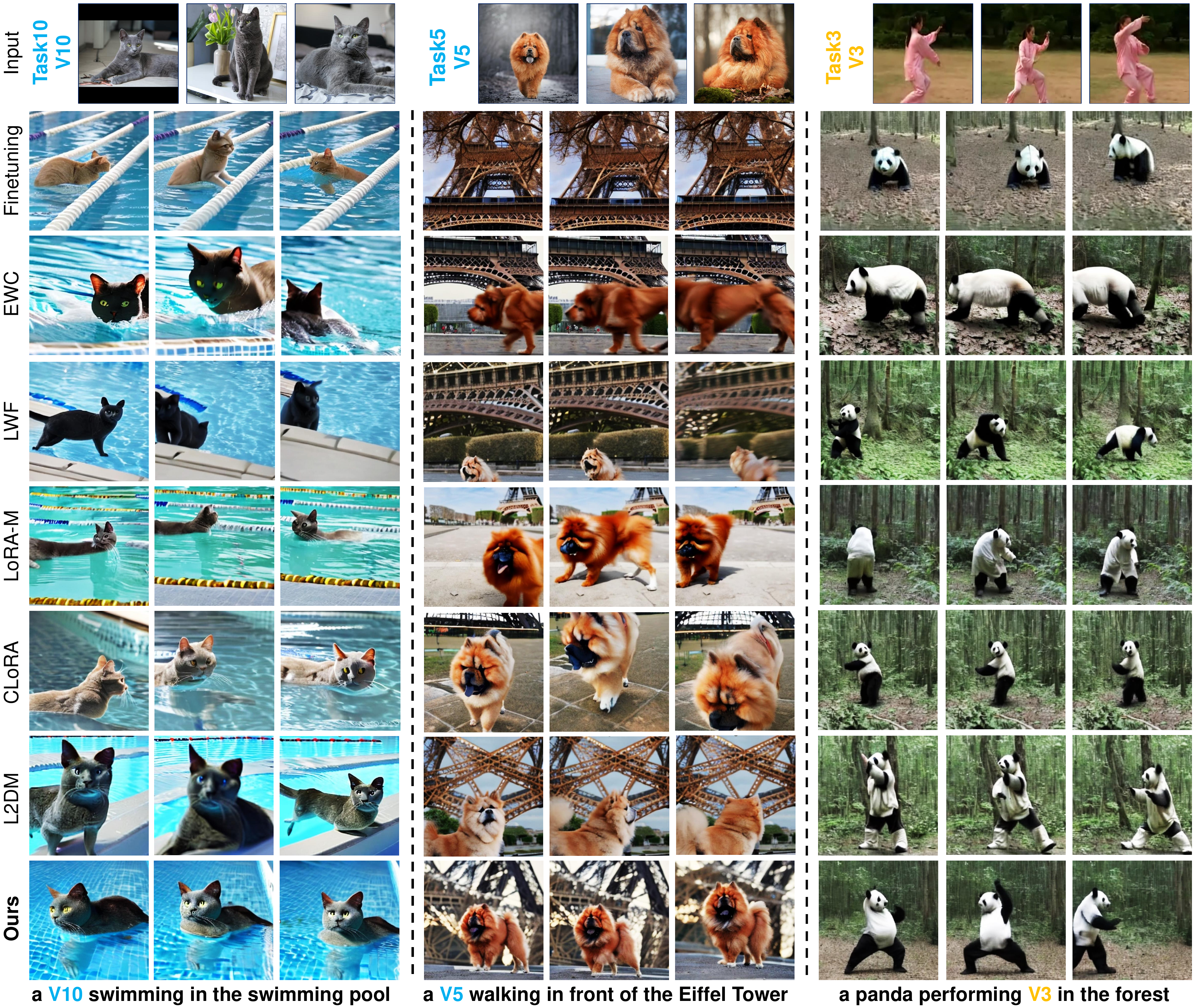}
\vspace{-7mm}
\caption{Comparison results of single-concept video customization under the CTVC setting when the backbone is DreamVideo. }
\label{fig: compar_single_concept_supp}
\end{figure*}

\begin{figure*}
\centering
\includegraphics[width =1.0\linewidth]
{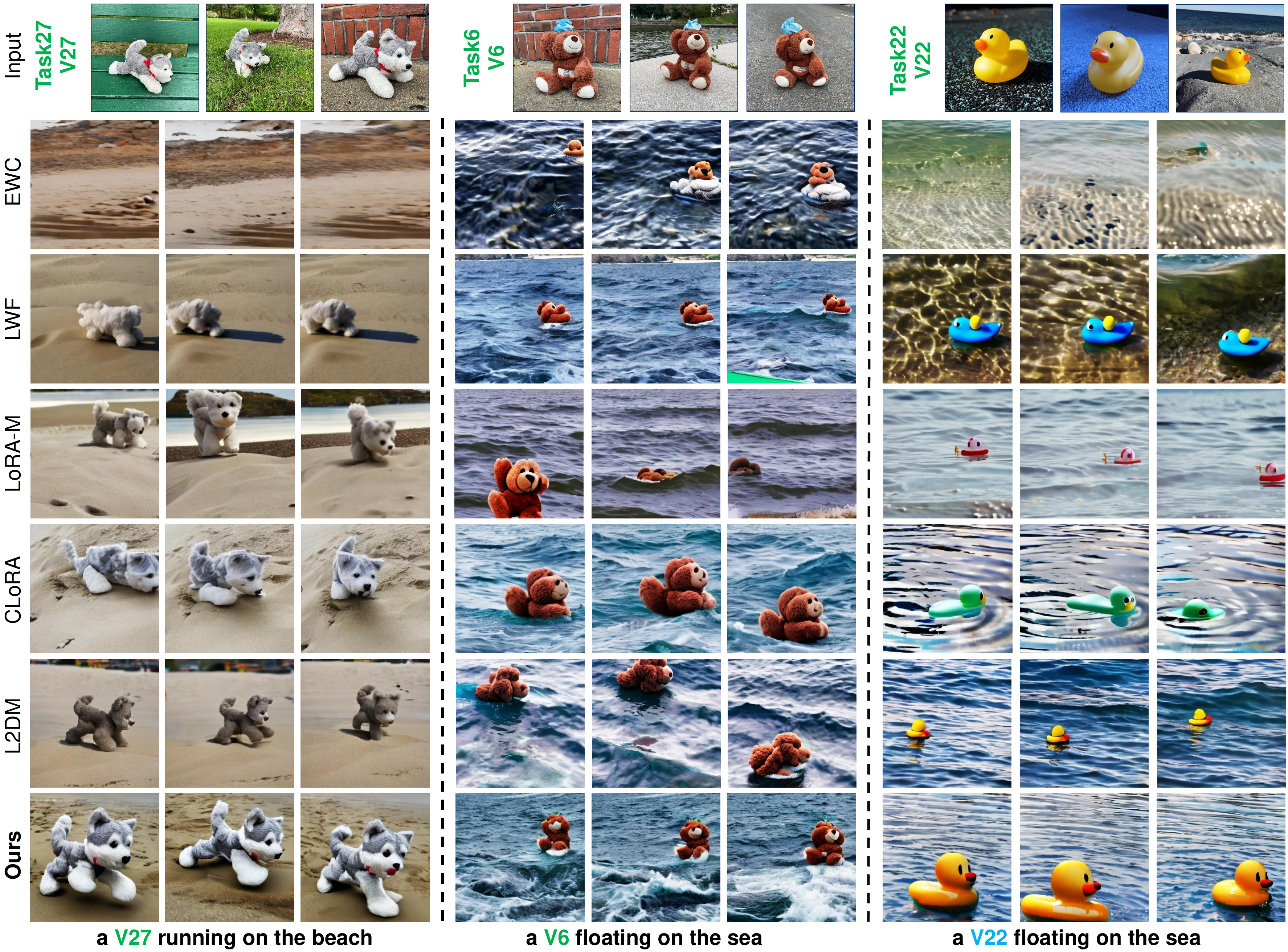}
\vspace{-7mm}
\caption{Comparison results of single-concept video customization under the CTVC setting when the backbone is DreamVideo. }
\label{fig: compar_single_concept_supp2}
\end{figure*}

\subsection{Qualitative Comparisons}
\textbf{Single-Concept Video Customization:}
As presented in Figs.~\ref{fig: compar_single_concept_supp}--\ref{fig: compar_single_concept_supp2}, we perform in-depth qualitative evaluations of single-concept video customization. In Fig.~\ref{fig: compar_single_concept_supp}, when provided with a text prompt (e.g., ``a panda performing V3 in the forest'') for inference, our model accurately adheres to the prompt instructions, producing videos that maintain the distinctive identity of the personalized concept V3. In comparison, existing comparison methods experience severe catastrophic forgetting, resulting in subpar videos with distorted motions and inconsistent identity representation.

\begin{figure*}
\centering
\includegraphics[width =1.0\linewidth]
{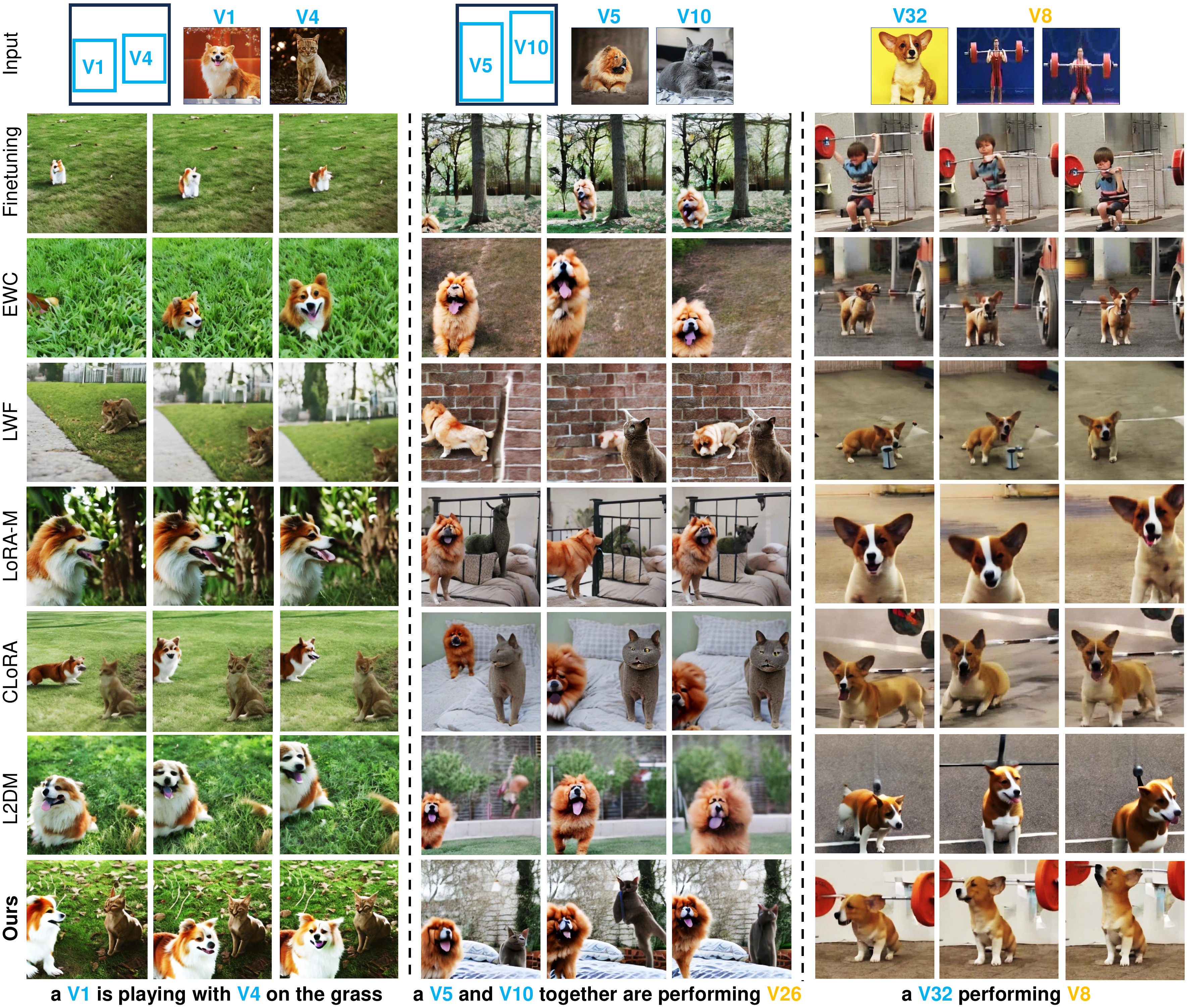}
\vspace{-7mm}
\caption{Comparison results of multi-concept video customization under the CTVC setting when the backbone is DreamVideo. }
\label{fig: compar_multi_concept_supp}
\end{figure*}

\textbf{Multi-Concept Video Customization:}
Fig.~\ref{fig: compar_multi_concept_supp} illustrates detailed comparative experiments on multi-concept video customization. To ensure fairness in evaluating against existing approaches, we incorporate the region-aware cross-attention mechanism from Mix-of-Show \cite{gu2023mixofshow} into the spatial transformer block (STB) of all SOTA competing methods. The qualitative analysis in Fig.~\ref{fig: compar_multi_concept_supp} highlights the exceptional capability of our model in multi-concept video customization, especially when continuously acquiring new personalized concepts under the CTVC setting.
Specifically, when provided with an initial text prompt and several region text prompts accompanied by user-defined bounding boxes, all competing methods display substantial concept neglect and struggle to achieve effective multi-concept video customization. Firstly, it is clear that these methods fail to retain the identity of previously learned concepts due to catastrophic forgetting. Secondly, they often overlook crucial subject concepts and motion patterns during the video generation process. For instance, while users might aim to synthesize a video featuring V5 and V10 performing V26 together, none of the existing comparative methods successfully generate both V10 and V26, as depicted in Fig.~\ref{fig: compar_multi_concept_supp}.

\begin{figure}
\centering
\includegraphics[width =1.0\linewidth]
{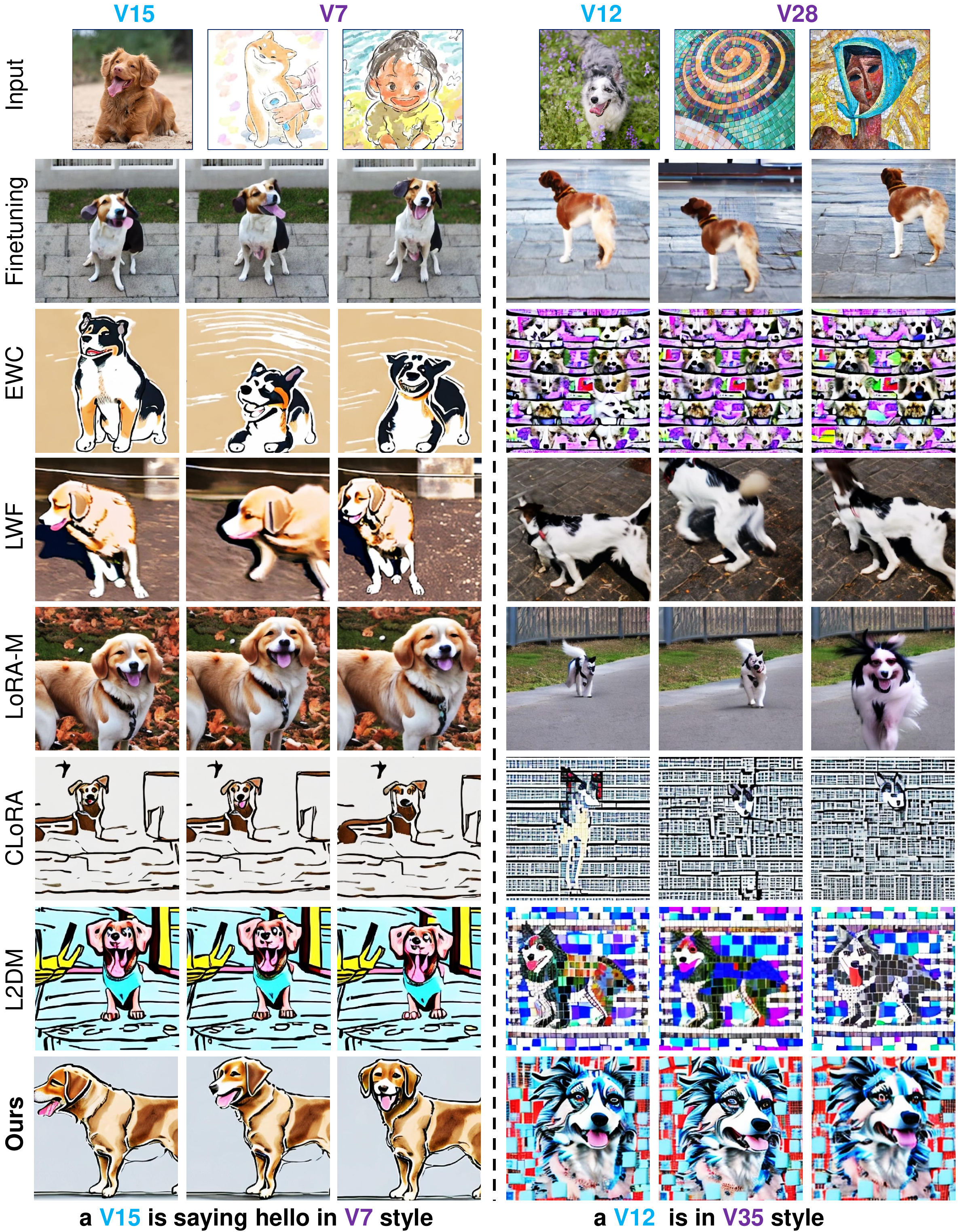}
\vspace{-7mm}
\caption{Comparison results of video style transfer under the CTVC setting when the backbone is DreamVideo. }
\label{fig: compar_style_transfer_supp}
\end{figure}

\textbf{Style Transfer:}
As illustrated in Fig.~\ref{fig: compar_style_transfer_supp}, we present comparative analyses of customized video style transfer to assess the efficacy of our proposed CCVD in tackling the challenges of the CTVC problem. From the results in Fig.~\ref{fig: compar_style_transfer_supp}, it is evident that our proposed model outperforms other existing methods in generating videos that align with user-provided style concepts. This demonstrates the critical role of the designed task-aware concept aggregation module in maintaining unique style attributes under the CTVC setting. In contrast, other competing methods face difficulties in capturing the identity of diverse style concepts, primarily due to the challenges of catastrophic forgetting and concept neglect in the CTVC setting.

\subsection{Societal Impact and Limitations}
\textbf{Societal Impact:}
Continual text-to-video customization (CTVC) has the potential to transform content creation by making high-quality personalized video production more accessible. This technology benefits various industries, including education, entertainment, marketing, and training, by allowing users to generate videos tailored to their specific needs. For educators, it enables the dynamic creation of instructional materials that adapt to different learning styles, making education more engaging and personalized. In entertainment, independent creators and small businesses can leverage this tool to produce unique, professional-grade videos without requiring expensive equipment or specialized expertise.
Furthermore, continual learning capabilities allow models to refine and expand their knowledge over time. Users can introduce new subjects, motion patterns, or artistic styles while maintaining previously learned concepts, resulting in a more flexible and adaptive video generation process. This personalization enhances user experience, enabling iterative improvements based on real-time feedback.

\textbf{Limitations:}
While CTVC offers numerous advantages, it also has some limitations. One is copyright infringement, as users might finetune models on proprietary materials without authorization. Another is privacy concerns, as models may require storing user preferences and personal data for customization. Balancing flexibility and privacy protection is crucial for the long-term success of this technology.

\end{document}